\documentclass{article}
\usepackage[dvipsnames, x11names]{xcolor}         % colors

\usepackage{hyperref}
\usepackage[accepted]{icml2023}

\usepackage{xspace}
\usepackage[utf8]{inputenc} % allow utf-8 input
\usepackage[T1]{fontenc}    % use 8-bit T1 fonts
\usepackage{url, color}            % simple URL typesetting
\usepackage{nicefrac}       % compact symbols for 1/2, etc.

\usepackage{microtype}
\usepackage{graphicx}
\usepackage{booktabs} % for professional tables
\usepackage{lscape}
\usepackage{rotating}
\usepackage[ruled, vlined]{algorithm2e}
\usepackage[utf8]{inputenc}
\usepackage{microtype}
\usepackage{graphicx}
\usepackage{float}
\usepackage{enumerate}
\usepackage{amsthm, amsmath, mathtools, amsfonts, amssymb, dsfont, enumitem, mathabx}
\usepackage{bm}
\usepackage{bbm}
\usepackage[mathscr]{euscript}
\usepackage{adjustbox}
\usepackage{footmisc}
\usepackage[capitalize,noabbrev]{cleveref}
\usepackage{multirow}
\usepackage{transparent}
\usepackage{caption}
\usepackage{subcaption}
\usepackage{bbm}

\newcommand{\Tau}{\mathcal{T}}
\newcommand{\TildeTau}{\widetilde{\mathcal{T}}}
\newcommand{\offlinedata}{\mathscr{T}_\text{offline}}

\newcommand{\set}[1]{\left\{#1\right\}}

\renewcommand{\hat}{\widehat}
\newcommand{\rstar}{r^\star}
\newcommand{\R}{\mathbb{R}}

\newcommand{\A}{\mathcal{A}}
\newcommand{\M}{\mathcal{M}}
\renewcommand{\S}{\mathcal{S}}
\renewcommand{\P}{\operatorname{\mathbb{P}}}
\newcommand{\E}{\operatorname{\mathbb{E}}}

\newcommand{\hopper}{\texttt{hopper}\xspace}
\newcommand{\walker}{\texttt{walker2d}\xspace}
\newcommand{\cheetah}{\texttt{halfcheetah}\xspace}

\newcommand{\medium}{\texttt{medium}\xspace}
\newcommand{\medreplay}{\texttt{med-replay}\xspace}
\newcommand{\medexpert}{\texttt{med-expert}\xspace}
\newcommand{\umaze}{\texttt{umaze}\xspace}
\newcommand{\mazemedium}{\texttt{medium}\xspace}
\newcommand{\mazelarge}{\texttt{large}\xspace}
\newcommand{\vzero}{\texttt{v0}\xspace}
\newcommand{\play}{\texttt{play}\xspace}
\newcommand{\diverse}{\texttt{diverse}\xspace}

\renewcommand{\epsilon}{\varepsilon}
\newcommand{\eps}{\varepsilon}
\newcommand{\wh}{\widehat}
\definecolor{blond}{rgb}{0.98, 0.94, 0.75}
\newcommand{\by}[1]{\colorbox{blond}{#1}}

\newcommand{\name}{CWBC\xspace}
\graphicspath{{./figure/}}

\theoremstyle{plain}

\theoremstyle{definition}

\theoremstyle{remark}

\usepackage[textsize=tiny]{todonotes}

\usepackage{titlesec,comment}

\icmltitlerunning{Reliable Conditioning of Behavioral Cloning for Offline Reinforcement Learning}

\begin{document}

\twocolumn[
\icmltitle{Reliable Conditioning of Behavioral Cloning \\for Offline Reinforcement Learning}



\icmlsetsymbol{equal}{*}

\begin{icmlauthorlist}
\icmlauthor{Tung Nguyen}{1}
\icmlauthor{Qinqing Zheng}{2}
\icmlauthor{Aditya Grover}{1}
\end{icmlauthorlist}

\icmlaffiliation{1}{UCLA}
\icmlaffiliation{2}{Meta AI Research}

\icmlcorrespondingauthor{Tung Nguyen}{tungnd@cs.ucla.edu}

\icmlkeywords{Machine Learning, ICML}

\vskip 0.3in
]



\printAffiliationsAndNotice{}  % leave blank if no need to mention equal contribution

\begin{abstract}
% The goal of offline reinforcement learning (RL) is to learn near-optimal policies from static logged datasets, thus sidestepping expensive online interactions.
Behavioral cloning (BC) provides a straightforward solution to offline RL by mimicking offline trajectories via supervised learning. Recent advances~\cite{chen2021decision, janner2021offline, emmons2021rvs} have shown that by conditioning on desired future returns, BC can perform competitively to their value-based counterparts, while enjoying much more simplicity and training stability. While promising, we show that these methods can be \emph{unreliable}, as their performance may degrade significantly when conditioned on high, out-of-distribution (ood) returns. This is crucial in practice, as we often expect the policy to perform better than the offline dataset by conditioning on an ood value. We show that this unreliability arises from both the suboptimality of training data and model architectures. We propose ConserWeightive Behavioral Cloning (\name), a simple and effective method for improving the reliability of conditional BC with two key components: trajectory weighting and conservative regularization.  Trajectory weighting upweights the high-return trajectories to reduce the train-test gap for BC methods, while conservative regularizer  encourages the policy to stay close to the data distribution for ood conditioning. We study CWBC in the context of RvS~\cite{emmons2021rvs} and Decision Transformers~\citep{chen2021decision}, and show that CWBC significantly boosts their performance on various benchmarks.
\end{abstract}

\section{Introduction}
\label{sec:intro}
% \begin{figure*}[t]
%     \centering
%     \includegraphics[width=0.95\textwidth]{figure/motivation_figure.pdf}
%     \caption{Illustrative figures demonstrating three possible scenarios for conditioning of BC methods for offline RL. The ideal scenario (\textbf{left}) is hard or even impossible to achieve with suboptimal offline data. 
%     On the other hand, return-conditioned RL methods can show unreliable generalization (\textbf{middle}), where the performance drops quickly after a certain point in the vicinity of the dataset maximum.
%     Our goal is to ensure reliable generalization (\textbf{right}) even when conditioned on ood returns. \ag{add legend for green and orange curves. if poossible break figure 1 into 3 subfigures and add Ideal, Unreliable, Reliable in sub-caption for better readability}\ag{also, a little weird that x and y axis have different scales}\ag{"actual returns" is informal, use "achieved returns"}}
%     \label{fig:motivation}
% \end{figure*}
\begin{figure*}[t]
     \centering
     \begin{subfigure}[b]{0.27\textwidth}
         \centering
         \includegraphics[width=\textwidth]{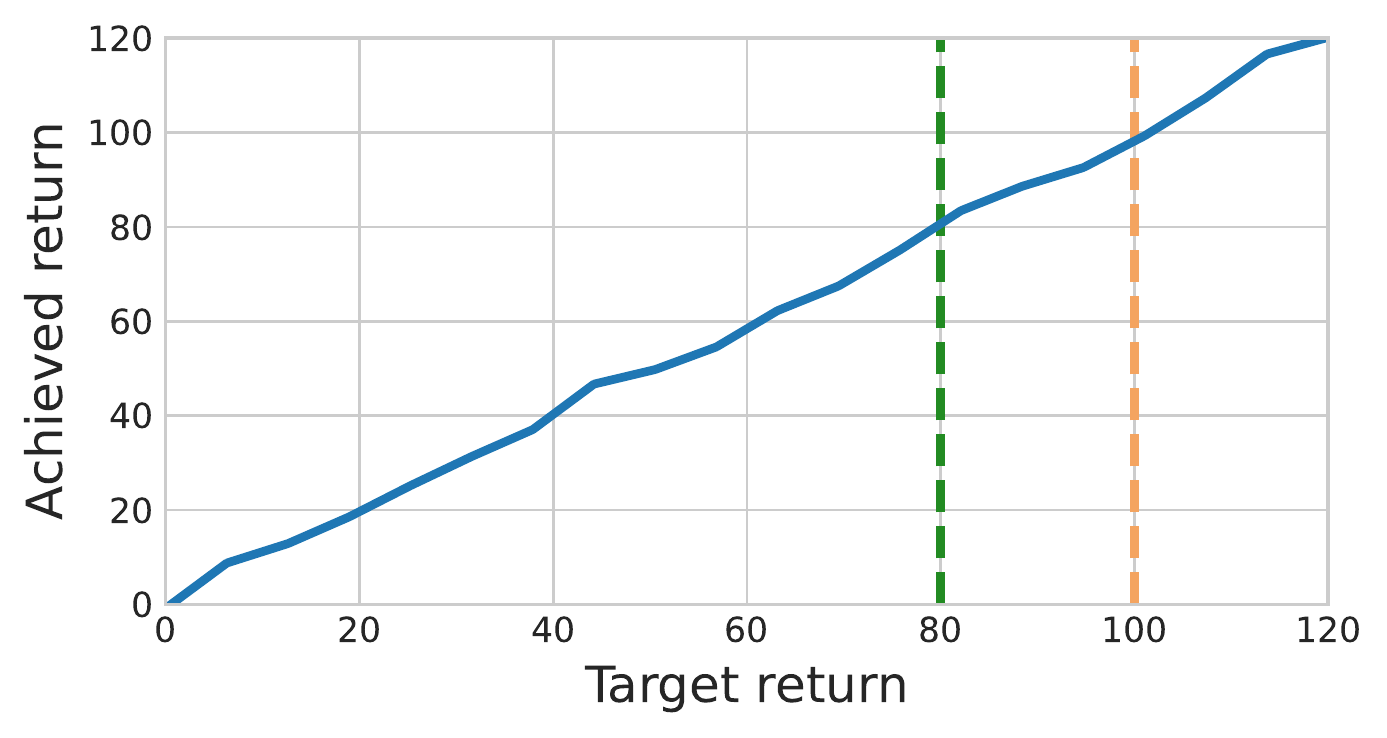}
         \caption{Ideal}
         \label{fig:ideal}
     \end{subfigure}
     % \hspace{20pt}
     \begin{subfigure}[b]{0.27\textwidth}
         \centering
         \includegraphics[width=\textwidth]{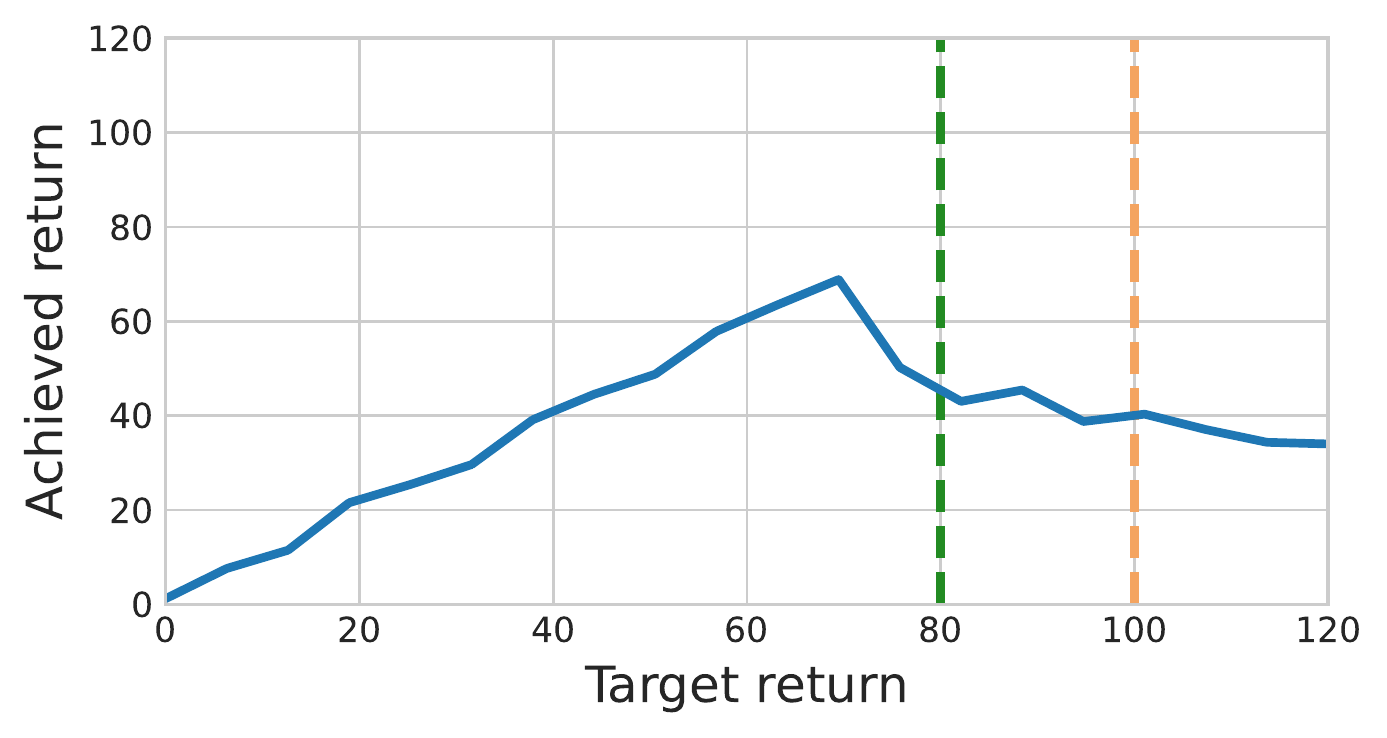}
         \caption{Unreliable}
        \label{fig:unreliable}
     \end{subfigure}
     \begin{subfigure}[b]{0.27\textwidth}
         \centering
         \includegraphics[width=\textwidth]{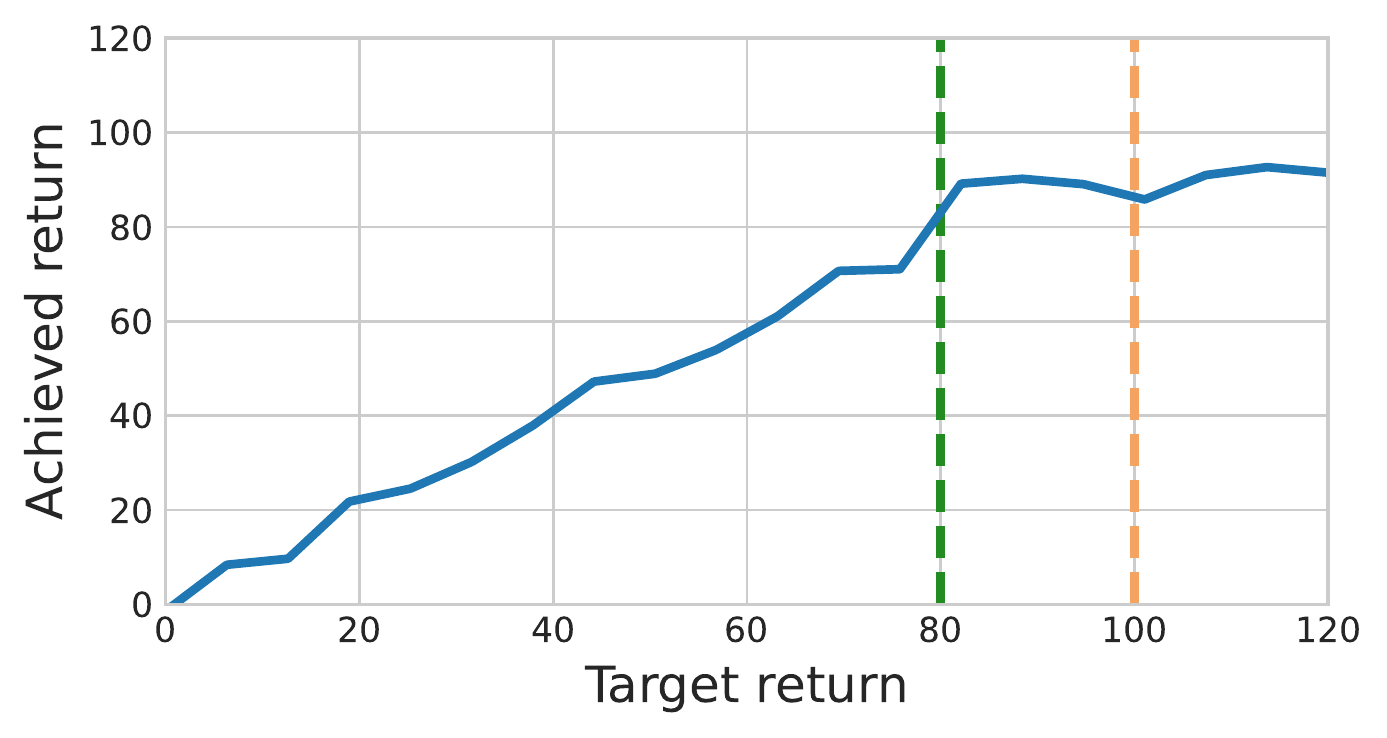}
         \caption{Reliable}
        \label{fig:reliable}
     \end{subfigure}
    \caption{Illustrative figures demonstrating three hypothetical scenarios for conditioning of BC methods for offline RL. The green line shows the maximum return in the offline dataset, while the orange line shows the expert return. The ideal scenario (a) is hard or even impossible to achieve with suboptimal offline data. 
    On the other hand, return-conditioned RL methods can show unreliable generalization (b), where the performance drops quickly after a certain point in the vicinity of the dataset maximum.
    Our goal is to ensure reliable generalization (c) even when conditioned on ood returns.
    % \tn{low priority TODO: make the curves thicker} \qq{also larger fontsize}\ag{low priority: can also have legend only for middle figure since it's the same for all 3 cases. that might mess up the scaling of middle figure vs rest. if so, one alternative is to just mention what green and orange are in the caption}
    }
    \label{fig:motivation}
\end{figure*}
In many real-world applications such as education, healthcare and autonomous driving, collecting data via online interactions is expensive or even dangerous.
However, we often have access to logged datasets in these domains that have been collected previously by some unknown policies.
The goal of offline reinforcement learning (RL) is to directly learn effective agent policies from such datasets, without additional online interactions~\cite{lange2012batch,levine2020offline}. 
Many online RL algorithms have been adapted to work in the offline setting, including value-based methods 
\cite{fujimoto2019off,ghasemipour2021emaq,wu2019behavior,jaques2019way,kumar2020conservative,fujimoto2021minimalist,kostrikov2021offline} as well as model-based methods~\cite{yu2020mopo,kidambi2020morel}. The key challenge in all these methods is to generalize the value or dynamics to state-action pairs outside the offline dataset.

An alternative way to approach offline RL is via approaches derived from behavioral cloning (BC)~\cite{bain1995framework}. BC is a supervised learning technique that was initially developed for imitation learning,
where the goal is to learn a policy that mimics expert demonstrations.
Recently, a number of works propose to formulate offline RL as supervised learning problems~\cite{chen2021decision,janner2021offline,emmons2021rvs}.
Since offline RL datasets usually do not have expert demonstrations,
these works condition BC on extra context information to specify target outcomes such as returns and goals. 
Compared with the value-based approaches, the empirical evidence has shown that these conditional BC approaches perform competitively, and they additionally enjoy the enhanced simplicity and training stability of supervised learning. 

As the maximum return in the offline trajectories is often far below the desired expert returns, we expect the policy to extrapolate over the offline data by conditioning on out-of-distribution (ood) expert returns. In an ideal world, the policy will achieve the desired outcomes, even when they are unseen during training. This corresponds to Figure~\ref{fig:ideal}, where the relationship between the achieved and target returns forms a straight line. In reality, however, the performance of current methods is far from ideal. Specifically, the actual performance closely follows the target return and peaks at a point near the maximum return in the dataset, but drops vastly if conditioned on a return beyond that point. Figure~\ref{fig:unreliable} illustrates this problem.

We systematically analyze the unreliability of current methods, and show that it depends on both the quality of offline data and the architecture of the return-conditioned policy. For the former, we observe that offline datasets are generally suboptimal and even in the range of observed returns, the distribution is highly non-uniform and concentrated over trajectories with low returns. 
This affects reliability, as we are mostly concerned with conditioning the policy on returns near or above the observed maximum in the offline dataset. One trivial solution to this problem is to simply filter the low-return trajectories prior to learning. However, this is not always viable as filtering can  eliminate a good fraction of the offline trajectories leading to poor data efficiency.

On the architecture aspect, we find that existing BC methods have significantly different behaviors when conditioning on ood returns. While DT~\citep{chen2021decision} generalizes to ood returns reliably, RvS~\citep{emmons2021rvs} is highly sensitive to such ood conditioning and exhibits vast drops in peak performance for such ood inputs. Therefore, the current practice for setting the conditioning return at test time in RvS is based on careful tuning with online rollouts, which is often tedious,  impractical, and inconsistent with the promise of offline RL to minimize online interactions.

While the idealized scenario in Figure~\ref{fig:ideal} is hard to achieve or even impossible depending on the training dataset and environment~\cite{wang2020statistical,zanette2021exponential,foster2021offline}, the unreliability of these methods is a major barrier for high-stakes deployments.
Hence, we focus this work on improving the reliability of return-conditioned offline RL methods. Figure~\ref{fig:reliable} illustrates this goal, where conditioning beyond the dataset maximum return does not degrade the model performance, even if the achieved returns do not match the target conditioning.
To this end, we propose ConserWeightive Behavior Cloning (\name{}), which consists of 2 key components: trajectory weighting and conservative regularization.
Trajectory weighting assigns and adjusts weights to each trajectory during training and prioritizes high-return trajectories for improved reliability.
Next, we introduce a notion of \emph{conservatism} for ood sensitive BC methods such as RvS, which encourages the policy to stay close to the observed state-action distribution when conditioning on high returns.
We achieve conservatism by selectively perturbing the returns of the high-return trajectories with a novel noise model and projecting the predicted actions to the ones observed in the unperturbed trajectory.
% We take trajectories with high returns from the dataset and add positive noise to their returns, which generates trajectories with large ood returns. 
% We predict actions conditioning on the perturbed returns and project them to the original actions by penalizing the $\ell_2$ distance. By imposing such a regularizer, we can condition the policy on large, unseen target returns at test time, sidestepping tedious manual tuning and online interactions. 

Our proposed framework is simple and easy to implement.
% sidestepping  online interactions for tuning the return values for conditioning. 
Empirically, we instantiate CWBC in the context of RvS~\citep{emmons2021rvs} and DT~\citep{chen2021decision}, two state-of-the-art BC methods for offline RL. CWBC significantly improves the performance of RvS and DT in D4RL~\citep{fu2020d4rl} locomotion tasks by $18\%$ and $8\%$, respectively, without any hand-picking of the value of the conditioning returns at test time.
% We further evaluate CWBC on Atari games~\citep{bellemare2013arcade} and show that it boosts the performance of RvS by $72\%$.

% We conduct extensive experiments on the D4RL~\cite{fu2020d4rl} benchmark. We confirm that \name{} significantly boosts the performance and the stability of RvS by $18\%$, outperforming all prior BC-based approaches, as well as surpassing the hybrid TD3+BC~\cite{fujimoto2021minimalist} approach for the first time, without any hand picking of the value of the conditioning returns at test-time.

\section{Preliminaries}
\label{sec:prelim}
\begin{figure*}[t]
    \centering
    \includegraphics[width=0.85\textwidth]{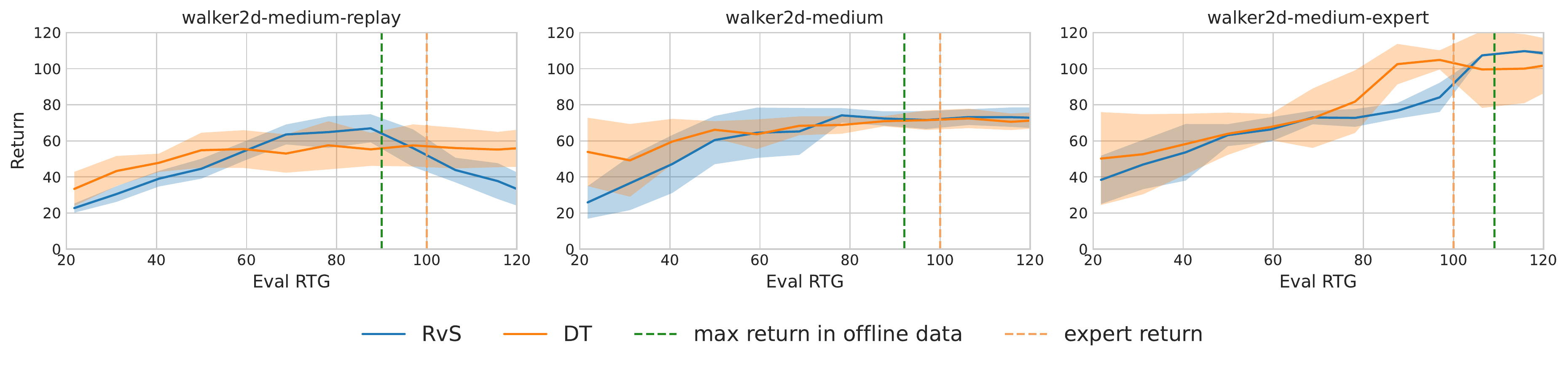}
    \includegraphics[width=0.85\textwidth]{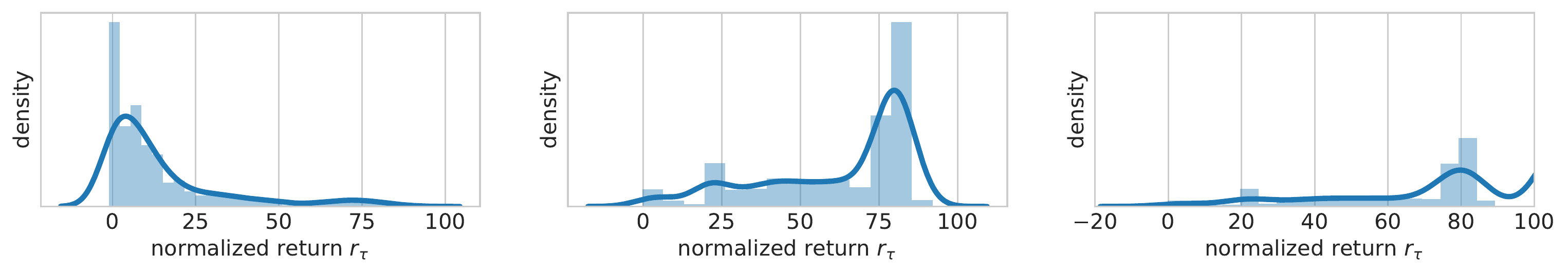}
    \caption{Reliability of RvS and DT on different \walker datasets. The first row shows the performance of the two methods, and the second row shows the return distribution of each dataset. Reliability decreases as the data quality decreases from \medexpert to \medreplay. While DT performs reliably, RvS exhibits vast drops in performance. 
    % \ag{put \medreplay first, then \med and \medexpert. improves readability}
    }
    \label{fig:reliability_problem}
\end{figure*}
We model our environment as a Markov decision process (MDP)~\citep{bellman1957mdp}, which can be described by a tuple  $\M= \langle\S, \A, p, P, R, \gamma \rangle$, where $\S$ is the state space, $\A$ is the action space, $p(s_1)$ is the distribution of the initial state, $P(s_{t+1}|s_t, a_t)$ is the transition probability distribution, $R(s_t, a_t)$ is the deterministic reward function, and $\gamma$ is the discount factor. At each timestep $t$, the agent observes a state $s_t \in S$ and takes an action $a_t \in \A$. This moves the agent to the next state $s_{t+1} \sim P(\cdot| s_t, a_t)$ and provides the agent with a reward $r_t = R(s_{t}, a_t)$.

\textbf{Offline RL.} We are interested in learning a (near-)optimal policy from a static offline dataset of trajectories collected by unknown policies, denoted as $\offlinedata$. We assume that these trajectories are \textit{i.i.d} samples drawn from some unknown static distribution $\Tau$. 
We use $\tau$ to denote a trajectory and use $|\tau|$ to denote its length.
Following ~\citet{chen2021decision}, the return-to-go (RTG) for a trajectory $\tau$ at timestep $t$ is defined as the sum of rewards starting from $t$ until the end of the trajectory: $g_t = \scriptstyle \sum_{t'=t}^{|\tau|} \textstyle r_{t'}$. This means the initial RTG $g_1$ is equal to the total return of the trajectory $r_\tau = \scriptstyle \sum_{t=1}^{|\tau|}\textstyle  r_t$.

\textbf{Decision Transformer (DT).} DT~\citep{chen2021decision} solves offline RL via sequence modeling. Specifically, DT employs a transformer architecture that generates actions given a sequence of historical states and RTGs. To do that, DT first transforms each trajectory in the dataset into a sequence of returns-to-go, states, and actions:
\begin{equation}
    \tau = \left(g_1, s_1, a_1, g_2, s_2, a_2, \dots, g_{|\tau|}, s_{|\tau|}, a_{|\tau|} \right).
\end{equation}
DT trains a policy that generates action $a_t$ at each timestep $t$ conditioned on the history of RTGs $g_{t-K:t}$, states $s_{t-K:t}$, and actions $a_{t-K:t-1}$, wherein $K$ is the context length of the transformer. The objective is a simple mean square error between the predicted actions and the ground truths:
\begin{equation}
\resizebox{\columnwidth}{!}{
    $\mathcal{L}_\text{DT}(\theta) = \E_{\tau \sim \Tau} \big[
    \textstyle \tfrac{1}{|\tau|}\sum_{t=1}^{|\tau|} \displaystyle \big(a_t - \pi_\theta(g_{t-K:t}, s_{t-K:t}, a_{t-K:t-1}) \big)^2 \big].$
    \label{eq:dt_loss}
}
\end{equation}
% }
During evaluation, DT starts with an initial state $s_1$ and a target RTG $g_1$. At each step $t$, the agent generates an action $a_t$, receives a reward $r_t$ and observes the next state $s_{t+1}$. DT updates its RTG $g_{t+1} = g_t - r_t$ and generates next action $a_{t+1}$. This process is repeated until the end of the episode.

\textbf{Reinforcement Learning via Supervised Learning (RvS).}
\citet{emmons2021rvs} conduct a thorough empirical study of conditional BC methods under the umbrella of Reinforcement Learning via Supervised Learning (RvS), and show that even simple models such as multi-layer perceptrons (MLP) can perform well. With carefully chosen architecture and hyperparameters, they exhibit performance that matches or exceeds the performance of transformer-based models.
There are two main differences between RvS and DT. First, RvS conditions on the average reward $\omega_t$ into the future instead of the sum of future rewards:
\begin{equation}
    \omega_t = \tfrac{1}{H - t + 1} \textstyle \sum_{t'=t}^{|\tau|} r_{t'} = \tfrac{g_t}{H - t + 1},
    \label{eq:condition_rvs}
\end{equation}
where $H$ is the maximum episode length. Intuitively, $\omega_t$ is RTG normalized by the number of remaining steps. Second, RvS employs a simple MLP architecture, which generates action $a_t$ at step $t$ based on only the current state $s_t$ and expected outcome $\omega_t$. RvS minimizes a mean square error:
\begin{equation}
    \mathcal{L}_\text{RvS}(\theta) = \E_{\tau \sim \Tau} \big[
    \textstyle \tfrac{1}{|\tau|}\sum_{t=1}^{|\tau|} \displaystyle \big(a_t - \pi_\theta(s_t, \omega_t) \big)^2 \big].
    \label{eq:obj_rvs}
\end{equation}
At evaluation, RvS is similar to DT, except that the expected outcome is now updated as $\omega_{t+1} = (g_t - r_t) / (H-t)$.
\section{Probing Unreliability of BC Methods} \label{sec:problem}
\begin{figure*}[t]
    \centering
    \includegraphics[width=0.85\textwidth]{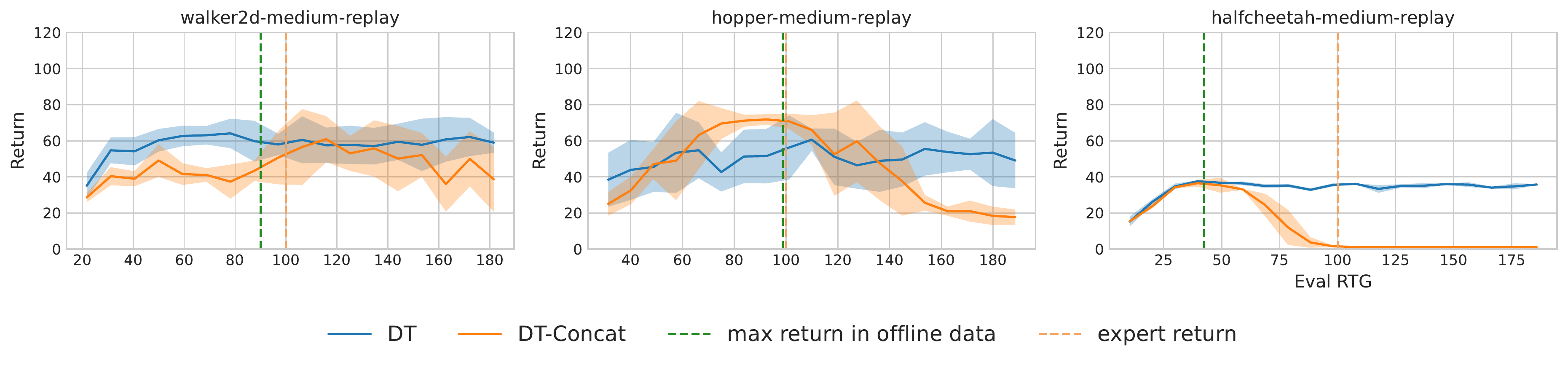}
    \caption{Performance of DT when the state and RTG tokens are concatenated. The results are averaged over $10$ seeds.}
    \label{fig:dt_concat}
\end{figure*}

Our first goal is to identify factors that influence the reliability of return-conditioned RL methods in practice.
To this end, we design 2 illustrative experiments distinguishing reliable and unreliable scenarios.

\textbf{Illustrative Exp 1 (Data centric ablation)}
In our first illustrative experiment, we show a run of RvS and DT on the \medreplay, \medium, and \medexpert datasets of the \walker environment from the D4RL~\citep{fu2020d4rl} benchmark. Figure~\ref{fig:reliability_problem} shows that reliability (top row) highly depends on the quality of the dataset (bottom row). 
Similar findings hold for other environments as well. 
In \medium and \medexpert datasets, RvS achieves a reliable performance when conditioned on high, out-of-distribution returns, while in the \medreplay dataset, the performance drops quickly after a certain point. This is because the \medreplay dataset has the lowest quality among the three, where most trajectories have low returns, as shown by the second row of Figure~\ref{fig:reliability_problem}. This low-quality data does not provide enough signal for the policy to learn to condition on high-value returns, thus negatively affecting the reliability of the model.

\textbf{Illustrative Exp 2 (Model centric ablation)}
Low-quality data is not the only cause of unreliability, but the architecture of the model also plays an important role. Figure~\ref{fig:reliability_problem} shows that unlike RvS, DT performs reliably in all three datasets. We hypothesize that the inherent reliability of DT comes from the transformer architecture. As the policy conditions on a sequence of both state tokens and RTG tokens to predict the next action, the attention layers can choose to ignore the ood RTG tokens while still obtaining a good prediction loss. In contrast, RvS employs an MLP architecture that takes both the current state and target return as inputs to generate actions, and thus cannot ignore the return information. To test this hypothesis, we experiment with a slightly modified version of DT, where we concatenate the state and RTG at each timestep instead of treating them as separate tokens. By doing this, the model cannot ignore the RTG information in the sequence. We call this version DT-Concat. Figure~\ref{fig:dt_concat} shows that the performance of DT-Concat is strongly correlated with the conditioning RTG, and degrades quickly when the target return is out-of-distribution. This result empirically confirms our hypothesis. 
\section{Conservative Behavioral Cloning With Trajectory Weighting}
\label{sec:algo}
We propose ConserWeightive Behavioral Cloning (\name), a simple but effective framework for improving the reliability of current BC methods. CWBC consists of two components, namely \emph{trajectory weighting} and \emph{conservative regularization}, which tackle the aforementioned issues relating to the observed data distribution and the choice of model architectures, respectively. Trajectory weighting provides a systematic way to transform the suboptimal data distribution to better estimate the optimal distribution by upweighting the high-return trajectories. Moreover, for BC methods such as RvS which use unreliable model parameterizations, we propose a novel conservative loss regularizer that encourages the policy to stay close to the data distribution when conditioned on large, ood returns. 
% We present the two components and their implementation details in the next two subsections.

\subsection{Trajectory Weighting} 
\label{subsec:algo_reweight}
To formalize our discussion, recall that $r_\tau$ denotes the return of a trajectory $\tau$ and let $\rstar = \sup_\tau r_\tau$ be the maximum expert return, which is assumed to be known in prior works on conditional BC~\cite{chen2021decision,emmons2021rvs}. We know that the optimal offline data distribution, denoted by \textbf{$\Tau^\star$}, is simply the distribution of demonstrations rolled out from the optimal policy.
Typically, the offline trajectory distribution $\Tau$ will be biased w.r.t. $\Tau^\star$.
During learning, this leads to a train-test gap, wherein we want to condition our BC agent on the expert returns during evaluation, but is forced to minimize the empirical risk on a biased data distribution during training.

The core idea of our approach is to transform $\Tau$ into a new distribution $\TildeTau$ that better estimates $\Tau^\star$. More concretely, $\TildeTau$ should concentrate on high-return trajectories, which mitigates the train-test gap.
One naive strategy is to simply filter out a small fraction of high-return trajectories from the offline dataset.
However, since we expect the original dataset to contain very few high-return trajectories, this will eliminate the majority of training data, leading to poor data efficiency. Instead, we propose to weight the trajectories based on their returns. 
Let $f_{\Tau}: \R \mapsto \R_+$ be the density function of $r_\tau$ where $\tau \sim \Tau$.
We consider the transformed distribution $\TildeTau$ whose density function $p_{\TildeTau}$ is
\vspace{-2pt}
\begin{equation}
    \label{eq:transformed_traj_density}
    p_{\TildeTau}(\tau) \ \propto \ \overbrace{\tfrac{f_{\Tau} (r_\tau)}{f_{\Tau}(r_\tau) + \lambda} \cdot \exp\big(-\tfrac{|r_\tau - r^{\star}|}{\kappa}\big)}^{\text{trajectory weight}}, 
\end{equation}
where $\lambda, \kappa \in \R_+$ are two hyperparameters that determine the shape of the transformed distribution. Specifically, $\kappa$ controls how much we want to favor the high-return trajectories, while $\lambda$ controls how close the transformed distribution is to the original distribution. Appendix~\ref{sec:kappa_lambda} provides a detailed analysis of the influence of these two hyperparameters on the transformed distribution.
% A larger value of $\kappa$ leads to a more uniform $\TildeTau$, whereas a smaller value upweights the high-return trajectories. In contrast, a smaller value of $\lambda$ gives more weights to high-return trajectories, while a larger value makes $\TildeTau$ closer to $\Tau$.
Our trajectory weighting is motivated by a similar scheme proposed for model-based design~\cite{kumar2020model}, where the authors use it to balance the bias and variance of gradient approximation for surrogates to black-box functions. 
We derive a similar theoretical result to this work in Appendix~\ref{sec:bias_variance}.
However, there are also notable differences. In model-based design, the environment is stateless and the dataset consists of $(x, y)$ pairs, whereas in offline RL we have a dataset of trajectories. Therefore, our trajectory weighting reweights the entire trajectories by their returns, as opposed to the original work which reweights individual data points. Moreover, the work of~\citet{kumar2020model} is based on GANs whereas we use vanilla supervised learning for our policy. 
% We adapt this weighting scheme to offline RL for trajectory sampling and discuss the implementation details below. 
\begin{figure}[t]
    \centering
    \includegraphics[width=0.8\columnwidth]{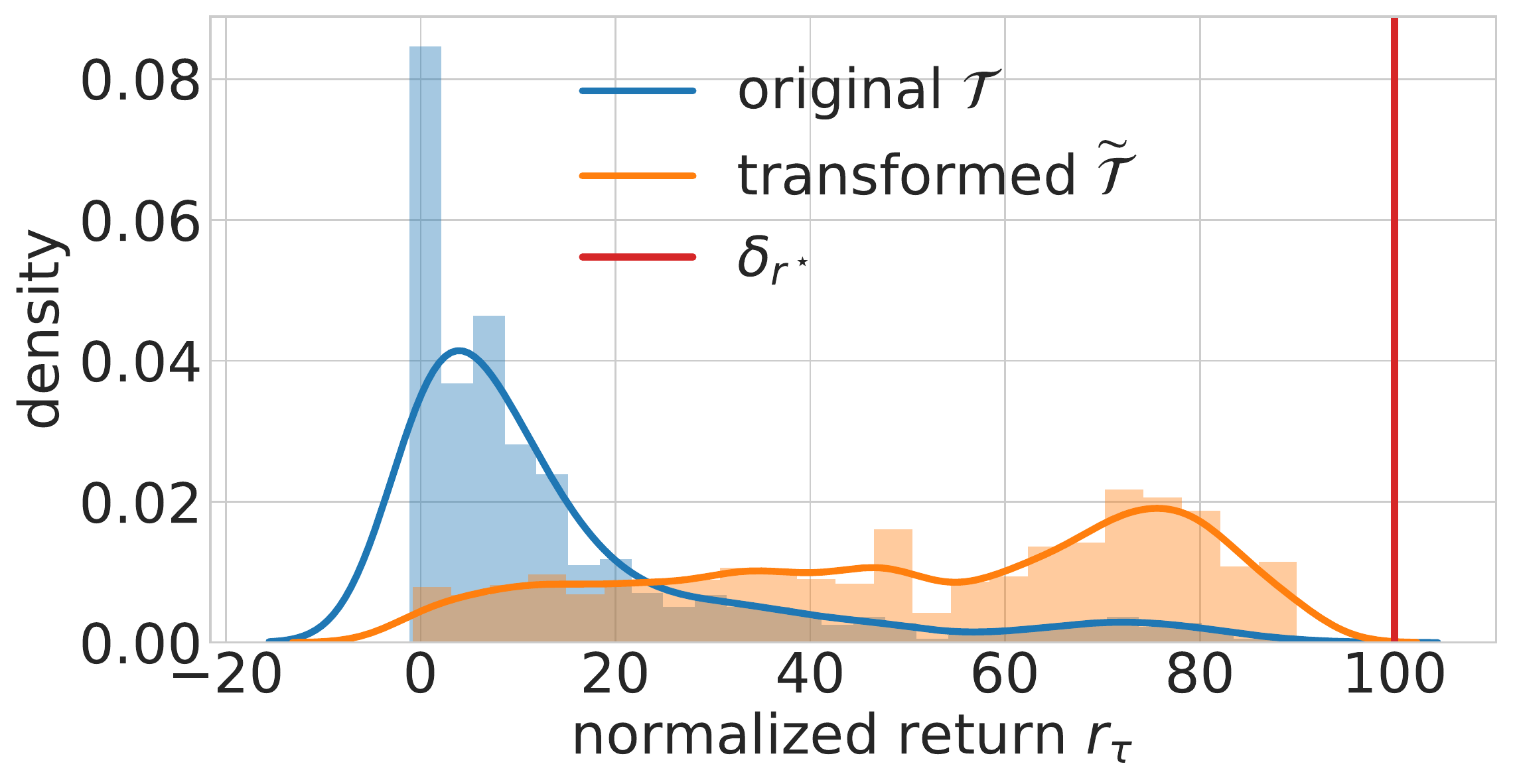}
    \caption{The original return distribution $\Tau$ and the transformed distribution $\TildeTau$ of \walker-\medreplay. We use $B=20$, $\lambda = 0.01$, $\kappa = \widehat{r}^\star - \wh{r}_{90}$, where $\wh{r}_{90}$ is the $90$-th percentile of the returns in the offline dataset.}
    \label{fig:transformation_density_walker}
\end{figure}

\subsubsection{Implementation Details} \label{sec:weighting_details}
In practice, the dataset $\offlinedata$ only contains a finite number of samples, and the density function $p_{\TildeTau}$ in equation~\eqref{eq:transformed_traj_density} cannot be computed exactly. Instead, we can
sample from a discretized approximation~\citep{kumar2020model} of $\TildeTau$.
We first group the trajectories in $\offlinedata$ into $B$ equal-sized bins according to the return $r_\tau$. To sample a trajectory, we first sample a bin index $b \in \set{1, \ldots, B}$ and then uniformly sample a trajectory inside bin $b$. We use $|b|$ to denote the size of bin $b$.
Let $\widebar{r}^b_{\tau} = 1/|b| \sum_{\tau \in b} r_{\tau}$ the average return of the trajectories in bin $b$,  $\widehat{r}^{\star}$ be the highest return in the dataset $\offlinedata$, and define $f_{\offlinedata}(b) = |b|/|\offlinedata| $.
As a discretized version of equation~\eqref{eq:transformed_traj_density}, the bins are weighted by their average returns with probability 
\begin{equation}
    \P_{\text{bin}}(b) \ \propto \ \tfrac{f_{\offlinedata}(b)}{f_{\offlinedata}(b) + \lambda} \cdot \exp \big(- \tfrac{|\widebar{r}^b_{\tau}- \widehat{r}^{\star}|}{\kappa} \big).
    \label{eq:bin_sampling_probability}
\end{equation}

 Figure~\ref{fig:transformation_density_walker} illustrates the impact of trajectory weighting on the return distribution of the \medreplay dataset for the \walker environment. We plot the histograms before and after transformation, where the density curves are estimated by kernel density estimators. 
 Algorithm~\ref{alg:weighted_sampling} summarizes the data sampling procedure with trajectory weighting.
\begin{algorithm}[t]
\small
 \DontPrintSemicolon
  \textbf{Input:} offline dataset $\offlinedata$, number of bins $B$, smoothing parameters $\lambda, \kappa$\;
  Compute the returns: $r_\tau \leftarrow \sum_{t=1}^{|\tau|} r_t, \; \forall \tau \in \offlinedata$.\;
  Group the trajectories into $B$ equal-sized bins according to $r_\tau$. \;
  Sample a bin $b \in [B]$ with probability $\P_\text{bin}(b)$ defined in Equation~\eqref{eq:bin_sampling_probability}.\;
  Sample a trajectory $\tau$ in bin $b$ uniformly at random. \;
  \textbf{Output:} $\tau$ 
  \caption{Weighted Trajectory Sampling}
\label{alg:weighted_sampling}
\end{algorithm}

\subsection{Conservative Regularization}
The architecture of the model also plays an important role in the reliability of BC methods. While the idealized scenario in Figure~\ref{fig:ideal} is hard or even impossible to achieve, we require a model to at least stay close to the data distribution to avoid catastrophic failure when conditioned on ood returns. In other words, we want the policy to be \emph{conservative}. Section~\ref{sec:problem} shows that DT enjoys self-conservatism, while RvS does not.
However, conservatism does not have to come from the architecture, but can also emerge from a proper objective function, as commonly done in conservative value-based methods~\citep{kumar2020conservative,fujimoto2021minimalist}. In this section, we propose a novel conservative regularization for return-conditioned BC methods that explicitly encourages the policy to stay close to the data distribution.
The intuition is to enforce the predicted actions when conditioning on large ood returns to stay close to the in-distribution actions. 
To do that, for a trajectory $\tau$ with high return, we inject \emph{positive} random noise $\eps \sim \mathcal{E}_\tau$ to its RTGs, and penalize the $\ell_2$ distance between the predicted action and the ground truth.
Specifically, to guarantee we generate large ood returns, we choose a noise distribution $\mathcal{E}$ such that the perturbed initial RTG $g_1 + \epsilon$ is at least $\hat{r}^\star$, the highest return in the dataset. The next subsections instantiate the conservative regularizer in the context of RvS.

\subsubsection{Implementation Details} \label{subsec:algo_conservatism}
We apply conservative regularization to trajectories whose returns are above $\wh{r}_{q}$, the $q$-th percentile of returns in the dataset. This makes sure that when conditioned on ood returns, the policy behaves similarly to high-return trajectories and not to a random trajectory in the dataset.
We sample a scalar noise $\eps \sim \mathcal{E}_\tau$ and offset the RTG of $\tau$ at every timestep by $\eps$: $g^\eps_t = g_t + \eps$, $t = 1, \ldots, |\tau|$, resulting in the conservative regularizer:
\begin{equation}
\resizebox{\columnwidth}{!}{
    $\mathcal{C}_\text{RvS}(\theta) = \E_{\tau \sim \Tau, \ \eps \sim \mathcal{E}_\tau} \big[ 
    \mathbbm{1}_{r_\tau > \wh{r}_{q}} \cdot \textstyle \tfrac{1}{|\tau|} \sum_{t=1}^{|\tau|} \displaystyle \big(a_t - \pi_\theta(s_t, \omega^\eps_t) \big)^2 \big],$
    \label{eq:conservatism_rvs}
}
\end{equation}
where $\omega^\eps_t = (g_t + \eps)/(H-t+1)$ (cf. Equation~\eqref{eq:condition_rvs}) is the noisy average RTG at timestep $t$. We observe that using the $95$-th percentile of $\wh{r}_{95}$ generally works well across different environments and datasets.
We use the noise distribution $\mathcal{E}_\tau = \text{Uniform}[l_\tau, u_\tau]$, where the lower bound
$l_\tau = \wh{r}^\star - r_\tau$ so that 
the perturbed initial RTG $g^\eps_1 = r_\tau + \eps$ is no less than $\wh{r}^\star$,
and the upper bound $u_\tau = \wh{r}^\star - r_\tau + \sqrt{12\sigma^2}$ so that
the standard deviation of $\mathcal{E}_\tau$ is equal to $\sigma$.
We emphasize our conservative regularizer is distinct from the other conservative components proposed for the value-based offline RL methods. While those usually try to regularize the value function estimation to prevent extrapolation error~\cite{fujimoto2019off}, we perturb the returns to generate ood conditioning and regularize the predicted actions. 

When the conservative regularizer is used, the final objective for training RvS is $\mathcal{L}_{\text{RvS}}(\theta) + \alpha \cdot \mathcal{C}_{\text{RvS}}(\theta)$, in which $\alpha$ is the regularization coefficient. When trajectory reweighting is used in conjunction with the conservative regularizer, we obtain \emph{ConserWeightive Behavioral Cloning (CWBC)}, which combines the best of both components. We provide a pseudo code for CWBC in Algorithm~\ref{alg:cwbc}.
\begin{algorithm}[t]
\small
 \DontPrintSemicolon
%  \small
  \textbf{Input:} dataset $\offlinedata$, number of iterations $I$, batch size $S$, regularization coefficient $\alpha$, initial parameters $\theta_0$ \\
%   Initialize policy parameters $\theta \leftarrow \theta_0$.\;
  \For{iteration $i = 1, \ldots, I$}{
    Sample a batch of trajectories $\mathcal{B} \leftarrow \{\tau^{(1)}, \ldots,\tau^{(S)} \}$ from $\offlinedata$ using Algorithm~\ref{alg:weighted_sampling}. \;
    %\ag{i think we should have steps here for how the perturbations are generated to evaluate the conservative reg}\\
    \For{every sampled trajectory $\tau^{(i)}$}{
        Samplie noise $\eps$ as described in Section~\ref{subsec:algo_conservatism}.\;
        Compute noisy RTGs: $g^\eps_t \leftarrow g_t + \eps$, $1 \leq t \leq |\tau^{(i)}|$.\;
    }
    % \tcp{loss and regularizer defined in Equation~\eqref{eq:obj_rvs} and~\eqref{eq:conservatism_rvs} }
    Perform gradient update of $\theta$ by minimizing the regularized empirical risk $\wh{\mathcal{L}}^\mathcal{B}_{\text{RvS}}(\theta) + \alpha \cdot \wh{\mathcal{C}}^\mathcal{B}_\text{RvS}(\theta)$. \;
  }
  \textbf{Output:} $\pi_\theta$ 
  \caption{ConserWeightive Behavioral Cloning (CWBC) for RvS}
\label{alg:cwbc}
\end{algorithm}

\section{Experiments} \label{sec:experiments}
\begin{table*}[t]
    \centering
    \caption{Comparison of the normalized return on the D4RL locomotion benchmark. For BC and TD3+BC, we get the numbers from~\cite{emmons2021rvs}. For IQL, we get the numbers from~\cite{iql}. For TTO, we get the numbers from~\citep{janner2021offline}. The results are averaged over 10 seeds.
    }
    \label{table:main_results}
    \resizebox{0.85\textwidth}{!}{
    % \begin{tabular}{l || l  l  l | l  l  l  | l  l  l | l | l}
    \begin{tabular}{l || l l l l l l l l}
    \toprule
    & RvS & RvS+CWBC & DT & DT+CWBC & TD3+BC & CQL & IQL & TTO \\ \midrule
    \walker-\medium & $73.3 \pm 5.7$ & \by{$73.6 \pm 5.4$} & $71.5 \pm 3.9$ & $70.4 \pm 4.5$ & $83.7$ & $82.9$ & $78.3$ & $81.3 \pm 8.0$ \\
    \walker-\medreplay & $54.0 \pm 12.1$ & \by{$72.8 \pm 7.5$} & $53.4 \pm 12.2$ & \by{$60.5 \pm 8.9$} & $81.8$ & $86.1$ & $73.9$ & $79.4 \pm 12.8$ \\
    \walker-\medexpert & $102.2 \pm 2.3$ & \by{$107.6 \pm 0.5$} & $99.8 \pm 21.3$ & \by{$108.2 \pm 0.8$} & $110.1$ & $109.5$ & $109.6$ & $91.0 \pm 10.8$ \\ \midrule
    \hopper-\medium & $56.6 \pm 5.5$ & \by{$62.9 \pm 3.6$} & $59.9 \pm 4.9$ & \by{$63.9 \pm 4.4$} & $59.3$ & $64.6$ & $66.3$ & $67.4 \pm 11.3$ \\
    \hopper-\medreplay & $87.7 \pm 9.7$ & $87.7 \pm 4.2$ & $56.4 \pm 20.1$ & \by{$76.9 \pm 5.9$} & $60.9$ & $97.8$ & $94.7$ & $99.4 \pm 12.6$ \\
    \hopper-\medexpert & $108.8 \pm 0.9$ & \by{$110.0 \pm 2.8$} & $95.4 \pm 11.3$ & \by{$103.4 \pm 9.0$} & $98.0$ & $102.0$ & $91.5$ & $106.0 \pm 1.1$ \\ \midrule
    \cheetah-\medium & $16.2 \pm 4.5$ & \by{$42.2 \pm 0.7$} & $42.5 \pm 0.6$ & $41.6 \pm 1.7$ & $48.3$ & $49.1$ & $47.4$ & $44.0 \pm 1.2$ \\
    \cheetah-\medreplay & $-0.4 \pm 2.7$ & \by{$40.4 \pm 0.8$} & $34.5 \pm 4.2$ & \by{$36.9 \pm 2.2$} & $44.6$ & $47.3$ & $44.2$ & $44.1 \pm 3.5$ \\
    \cheetah-\medexpert & $83.4 \pm 2.1$ & \by{$91.1 \pm 2.0$} &  $87.2 \pm 2.7$ & $85.6 \pm 2.0$ & $90.7$ & $85.8$ & $86.7$ & $40.8 \pm 8.7$ \\
    \midrule
    \# wins & / & $8$ & / & $6$ & / & / & / \\
    average & $64.6$ & $76.5$ & $66.7$ & $71.9$ & $75.3$ & $80.6$ & $77.0$ & $72.6$ \\
    \bottomrule
    \end{tabular}
    }
\end{table*}
The goal of our experiments is two-fold: (a) evaluate \name{} against existing approaches for offline RL on standard offline RL benchmarks, (b) ablate and analyze the individual contributions and interplay between trajectory weighting and conservative regularization. Additionally, we compare CWBC with max-return conditioning, a simple baseline for improving the reliability of return-conditioned methods.

\subsection{Evaluation on D4RL Benchmark}
\textbf{Dataset} We evaluate the effectiveness of CWBC on three locomotion tasks with dense rewards from the D4RL benchmark \cite{fu2020d4rl}: \hopper, \walker and \cheetah. For each task, we consider the v2 \medium, \medreplay and \medexpert offline datasets. The \medium dataset contains $1$M samples from a policy trained to approximately $\frac{1}{3}$ the performance of an expert policy. The \medreplay dataset uses the replay buffer of a policy trained up to the performance of a \medium policy. The \medexpert dataset contains $1$M samples generated by a \medium policy and $1$M samples generated by an expert policy.

\textbf{Baselines} We apply CWBC to RvS~\citep{emmons2021rvs} and DT~\citep{chen2021decision}, two state-of-the-art BC methods, which we denote as RvS-CWBC and DT-CWBC, respectively. In addition, we report the performance of TTO~\citep{janner2021offline} and three value-based methods: TD3+BC~\cite{fujimoto2021minimalist}, CQL~\cite{kumar2020conservative}, and IQL~\cite{iql} as a reference.

\textbf{Hyperparameters} We use a fixed set of hyperparameters of CWBC across all datasets. For trajectory weighting, we use $B=20$ and $\lambda = 0.01$, and we set the temperature parameter $\kappa$ to be the difference between the highest return and the $90$-th percentile: $\widehat{r}^{\star} - \wh{r}_{90}$, whose value varies across the datasets. For RvS, we apply our conservative regularization to trajectories whose returns are above the $q=95$-th percentile return in the dataset, and perturb their RTGs as described in Section~\ref{subsec:algo_conservatism}. We use a regularization coefficient of $\alpha=1$.
The model architecture and the other hyperparamters are identical to what were used in the original paper. We provide a complete list of hyperparameters in Appendix~\ref{sec:hyper} and additional ablation experiments on  $\lambda$ and $\kappa$ in Appendix~\ref{app:ablation}. At test time, we set the evaluation RTG to be the expert return for each environment.

\textbf{Results} Table~\ref{table:main_results} reports the performance of different methods we consider. DT+CWBC outperforms the original DT in $6/9$ datasets and performs comparably in others, achieving an average improvement of $8\%$. The improvement is significant in low-quality datasets (\medreplay), which is consistent with our analysis in Section~\ref{sec:problem}.

When applied to RvS, CWBC significantly improves the performance of the original RvS on $8/9$ datasets, while performing similarly in \hopper-\medreplay. RvS+CWBC achieves an average improvement of $18\%$ over RvS, making it the best performing BC method in the table. RvS+CWBC also performs competitively with the value-based methods, and only slightly underperforms CQL. We note that our reported numbers for RvS are different from the original paper~\citep{emmons2021rvs}, which is due to the difference in evaluation protocols. The original RvS paper tuned the target return for each environment, while we used expert return for all environments. The performance of RvS crashes when conditioning on this expert return, resulting in the difference in reported numbers. We can also tune the target return as done in the RvS paper, but this goes against the setting of offline RL that restricts online interactions.

\subsubsection{The Impact of Trajectory Weighting and Conservative Regularization}
\begin{figure*}[t]
    \centering
    \includegraphics[width=0.85\textwidth]{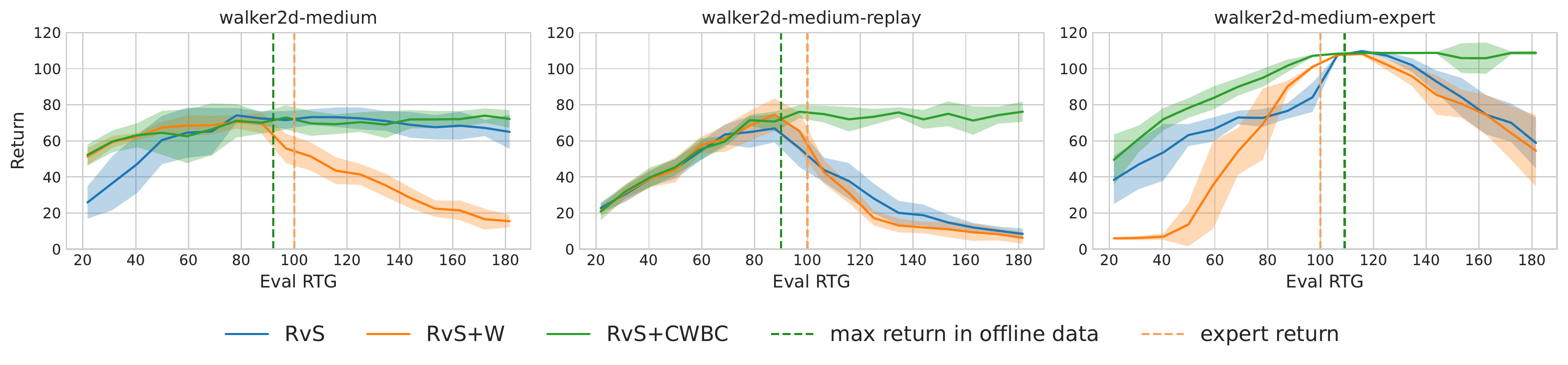}
    \caption{The performance of RvS and its two variants on \walker datasets. RvS+W denotes RvS with trajectory weighting only, while RvS+CWBC is RvS with both trajectory weighting and conservative regularization. 
    % Trajectory weighting has varying effects, but shows consistent performance improvements when used in conjunction with conservative regularization.
    }
    \label{fig:interplay}
\end{figure*}
To better understand the impact of each component in CWBC, we plot the achieved returns of RvS and its variants when conditioning on different target returns. Specifically, we compare RvS with RvS+W, a variant that only uses trajectory weighting, and RvS+CWBC with both trajectory weighting and conservative regularization enabled. Figure~\ref{fig:interplay} shows the performance of the three baselines. Trajectory weighting improves the performance of RvS when conditioning on high-value returns, especially in the low-quality \medreplay dataset. This agrees with our observation in Section~\ref{sec:problem}. However, trajectory weighting alone does not achieve reliability, as the performance still degrades quickly after the maximum return in the training data, leading to inconsistent improvements. 

When trajectory weighting is used in conjunction with conservative regularization, RvS+CWBC enjoys both better performance when conditioned on high returns, and better reliability with respect to out-of-distribution returns. This illustrates the significant importance of encouraging conservatism for RvS. By explicitly asking the model to stay close to the data distribution, we achieve more reliable out-of-distribution performance, and avoid the performance crash problem. This leads to absolute performance improvement of RvS+CWBC in Table \ref{table:main_results}.

\subsubsection{CWBC versus max-return conditioning} \label{sec:max_return_condition}
\begin{figure*}[t]
    \centering
    \includegraphics[width=0.85\textwidth]{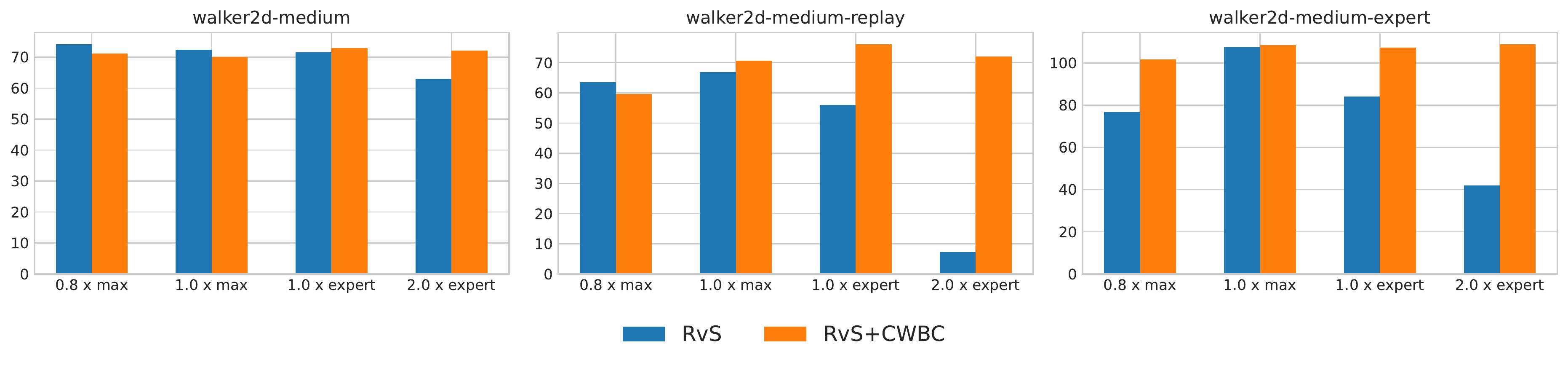}
    \caption{Performance of RvS and RvS+CWBC under different conditioning strategies.}
    \label{fig:max_return_condition}
\end{figure*}
One simple solution for the unreliability problem of return-conditioned methods is to map the target return to a fixed value whenever we condition the policy on an out-of-distribution return. As we have observed in Section~\ref{sec:problem}, this fixed value should be close to the maximum return in the offline dataset. Figure~\ref{fig:max_return_condition} compares the performance of RvS and RvS+CWBC when conditioning on either $0.8 \times$ max offline return, $1.0 \times$ max offline return, $1.0 \times$ expert return, or $2.0 \times$ expert return. The result shows that there is no common target return that achieves the best performance for RvS. In \walker-\medium, the best conditioning value is $0.8 \times$ maximum offline return, while it is $1.0 \times$ maximum offline return for \walker-\medreplay and \walker-\medexpert. The RvS paper also concluded that the best conditioning value is problem-specific, and had to tune this hyperparameter for each of the datasets.

By conditioning on the expert return, RvS+CWBC matches or even surpasses the best achieved performance of RvS across three datasets. This allows us to have a unified evaluation protocol for all tasks and datasets, sidestepping tedious and impractical finetuning of the conditioning parameter. Moreover, always clipping ood returns to the maximum return in the dataset impedes the model from extrapolating beyond offline data. In contrast, CWBC does not prohibit extrapolation. While conservative regularization encourages the policy to stay close to the data distribution, there is always a trade-off between optimizing the original supervised objective (which presumably allows extrapolation) and the conservative objective. Appendix~\ref{sec:atari_benchmark} provides additional experiments that show extrapolation behavior of CWBC.
\section{Related Work}
\textbf{Offline Temporal Difference Learning} Most of the existing off-policy RL methods are often based on temporal difference (TD) updates. A key challenge of directly applying them in the offline setting is the \emph{extrapolation error}: the value function is poorly estimated at unseen state-action pairs. To remedy this issue, various forms of \emph{conservatism} have been introduced to encourage the learned policy to stay close to the behavior policy that generates the data. 
For instance, \citet{fujimoto2019off,ghasemipour2021emaq} use certain policy parameterizations specifically tailored for offline RL. \citet{wu2019behavior, jaques2019way, kumar2019stabilizing} penalize the divergence-based distances between the learned policy and the behavior policy. \citet{fujimoto2021minimalist} propose an extra behavior cloning term to regularize the policy. 
This regularizer is simply the $\ell_2$ distance between predicted actions and the truth, yet surprisingly effective for porting off-policy TD methods to the offline setting. Instead of regularizing the policy, other works have sought to incorporate divergence regularizations into the value function estimation, e.g., \citep{nachum2019algaedice, kumar2020conservative, kostrikov2021offline}. Another recent work by \citet{iql} predicts the $Q$ function via expectile regression, where the estimation of the maximum $Q$-value is constrained to be in the dataset. 

\textbf{Behavior Cloning Approaches for Offline RL} 
Recently, there is a surge of interest in converting offline RL into supervised learning paradigms~\citep{chen2021decision, janner2021offline, emmons2021rvs}. In essence, these approaches conduct behavior cloning~\cite{bain1995framework} by additionally conditioning on extra information such as goals or rewards. Among these works, \citet{chen2021decision} and \citet{janner2021offline} have formulated offline RL as sequence modeling problems and train transformer architectures~\citep{vaswani2017attention} in a similar fashion to language and vision~\citep{radford2018improving,chen2020generative,brown2020language,lu2021pretrained, yan2021videogpt}. Extensions have also been proposed in the context of sequential decision making for offline black-box optimization~\cite{nguyen2022transformer,krishnamoorthy2022generative}. A recent work by~\citet{emmons2021rvs} further shows that conditional BC can achieve competitive performance even with a simple but carefully designed MLP network. 
Earlier, similar ideas have also been proposed for online RL, where the policy is trained via supervised learning techniques to fit the data stored in the replay buffer~\citep{schmidhuber2019reinforcement, srivastava2019training, ghosh2019learning}.

\textbf{Data Exploration for Offline RL} Recent research efforts have also been made towards understanding properties and limitations of datasets used for offline RL~\cite{yarats2022don, lambert2022challenges,guo2021learning}, particularly focusing on exploration techniques during data collection. 
Both \citet{yarats2022don} and \citet{lambert2022challenges} collect datasets using task-agnostic exploration strategies~\cite{laskin2021urlb}, relabel the rewards and train offline RL algorithms on them.  \citet{yarats2022don} benchmark multiple offline RL algorithms on different tasks including transferring, whereas \citet{lambert2022challenges} focus on improving the exploration method.

\section{Conclusion} \label{sec:conclusion}
We proposed ConserWeightive Behavioral Cloning (CWBC), a new framework that improves the reliability of BC methods in offline RL with two novel components: trajectory weighting and conservative regularization.
Trajectory weighting reduces the train-test gap when learning from a suboptimal dataset, improving the performance of both DT and RvS. Next, we showed that we achieve better reliability for ood sensitive methods such as RvS by using our proposed conservative regularizer. As confirmed by the experiments, CWBC significantly improves the performance and stability of RvS.
% CWBC balances the bias-variance tradeoff that arises in learning from a suboptimal dataset via trajectory weighting, and improves out-of-distribution conditioning based on a novel notion of action conservatism
% for offline RL.\ag{i think we should split the contributions and not talk about cwbc as a whole. traj weighting does xxx and improves both dt and rvs. next, we showed that while dt is self-conservative due to attention, we can recover this property even for rvs using ...}
% % Trajectory weighting shifts the offline data distribution towards the high-return trajectories (reduce bias), while also seek to control the increase in variance.
% % The conservative regularization stabilizes the performance of BC when conditioning on large out-of-distribution RTGs, which eases the hurdle of manually tuning the conditioning parameter.
% Confirmed by the experiments, CWBC substantially improves the performance of RvS and DT, two state-of-the-art  BC algorithms for offline RL. \ag{i think we need to be more precise here. only weighting improves DT }
% While empirically appealing, our proposed framework lacks a strong theoretical backup. 

While we made good progress for BC, advanced value-based methods such as CQL and IQL are still ahead and we believe a further understanding of the tradeoffs in both kinds of approaches is important for future work. Another promising direction from a data perspective is how to combine datasets from multiple environments to obtain diverse, high-quality data. Recent works have shown promising results in this direction~\cite{reed2022generalist}. Last but not least, while CWBC significantly improves the reliability of RvS, it is not able to achieve the idealized scenario in Figure~\ref{fig:ideal}. How to obtain strong generalization beyond offline data, or whether it is possible, is still an open question, and poses a persistent research opportunity for not only CWBC but the whole offline RL community.

% \section*{Reproducibility Statement}
% We present the practical implementation of our framework in Section~\ref{sec:weighting_details} and Section~\ref{sec:algo}. We include the implementation details of our paper in Appendix~\ref{sec:appendix}, which contains information about the datasets we use, the open sourced code we base on, and the list of hyperparameters we use to reproduce our results. Finally, we submitted the source code in the supplementary material.
% Appealingly, both the transformation procedure and the conservatism regularizer are extremely simple and can be readily applied into other conditional BC algorithms, without compromising on their default stability and scalability. 

% For practical implementation, these two components can be seamlessly integrated into any conditional BC algorithms with only a few lines' modification of the code, requiring negligible engineering effort. 
% \qq{add future work}

% \section*{Acknowledgements}
% We would like to thank Hritik Bansal, Shashank Goel, Siddarth Krishnamoorthy, Satvik Mashkaria, and Tuan Pham for the insightful discussions during the early development of the paper. We also want to thank the Google TPU Research Cloud program for the computing resources.

\newpage
\bibliography{main}

\begin{thebibliography}{43}
\providecommand{\natexlab}[1]{#1}
\providecommand{\url}[1]{\texttt{#1}}
\expandafter\ifx\csname urlstyle\endcsname\relax
  \providecommand{\doi}[1]{doi: #1}\else
  \providecommand{\doi}{doi: \begingroup \urlstyle{rm}\Url}\fi

\bibitem[Agarwal et~al.(2020)Agarwal, Schuurmans, and
  Norouzi]{agarwal2020optimistic}
Agarwal, R., Schuurmans, D., and Norouzi, M.
\newblock An optimistic perspective on offline reinforcement learning.
\newblock In \emph{International Conference on Machine Learning}, pp.\
  104--114. PMLR, 2020.

\bibitem[Bain \& Sammut(1995)Bain and Sammut]{bain1995framework}
Bain, M. and Sammut, C.
\newblock A framework for behavioural cloning.
\newblock In \emph{Machine Intelligence 15}, pp.\  103--129, 1995.

\bibitem[Bellemare et~al.(2013)Bellemare, Naddaf, Veness, and
  Bowling]{bellemare2013arcade}
Bellemare, M.~G., Naddaf, Y., Veness, J., and Bowling, M.
\newblock The arcade learning environment: An evaluation platform for general
  agents.
\newblock \emph{Journal of Artificial Intelligence Research}, 47:\penalty0
  253--279, 2013.

\bibitem[Bellman(1957)]{bellman1957mdp}
Bellman, R.
\newblock A markovian decision process.
\newblock \emph{Indiana Univ. Math. J.}, 1957.

\bibitem[Brown et~al.(2020)Brown, Mann, Ryder, Subbiah, Kaplan, Dhariwal,
  Neelakantan, Shyam, Sastry, Askell, Agarwal, Herbert-Voss, Krueger, Henighan,
  Child, Ramesh, Ziegler, Wu, Winter, Hesse, Chen, Sigler, Litwin, Gray, Chess,
  Clark, Berner, McCandlish, Radford, Sutskever, and Amodei]{brown2020language}
Brown, T.~B., Mann, B., Ryder, N., Subbiah, M., Kaplan, J., Dhariwal, P.,
  Neelakantan, A., Shyam, P., Sastry, G., Askell, A., Agarwal, S.,
  Herbert-Voss, A., Krueger, G., Henighan, T., Child, R., Ramesh, A., Ziegler,
  D.~M., Wu, J., Winter, C., Hesse, C., Chen, M., Sigler, E., Litwin, M., Gray,
  S., Chess, B., Clark, J., Berner, C., McCandlish, S., Radford, A., Sutskever,
  I., and Amodei, D.
\newblock Language models are few-shot learners, 2020.

\bibitem[Chen et~al.(2021)Chen, Lu, Rajeswaran, Lee, Grover, Laskin, Abbeel,
  Srinivas, and Mordatch]{chen2021decision}
Chen, L., Lu, K., Rajeswaran, A., Lee, K., Grover, A., Laskin, M., Abbeel, P.,
  Srinivas, A., and Mordatch, I.
\newblock Decision transformer: Reinforcement learning via sequence modeling.
\newblock \emph{Advances in neural information processing systems}, 34, 2021.

\bibitem[Chen et~al.(2020)Chen, Radford, Child, Wu, Jun, Luan, and
  Sutskever]{chen2020generative}
Chen, M., Radford, A., Child, R., Wu, J., Jun, H., Luan, D., and Sutskever, I.
\newblock Generative pretraining from pixels.
\newblock In \emph{International Conference on Machine Learning}, pp.\
  1691--1703. PMLR, 2020.

\bibitem[Emmons et~al.(2021)Emmons, Eysenbach, Kostrikov, and
  Levine]{emmons2021rvs}
Emmons, S., Eysenbach, B., Kostrikov, I., and Levine, S.
\newblock Rvs: What is essential for offline rl via supervised learning?
\newblock \emph{arXiv preprint arXiv:2112.10751}, 2021.

\bibitem[Foster et~al.(2021)Foster, Krishnamurthy, Simchi-Levi, and
  Xu]{foster2021offline}
Foster, D.~J., Krishnamurthy, A., Simchi-Levi, D., and Xu, Y.
\newblock Offline reinforcement learning: Fundamental barriers for value
  function approximation.
\newblock \emph{arXiv preprint arXiv:2111.10919}, 2021.

\bibitem[Fu et~al.(2020)Fu, Kumar, Nachum, Tucker, and Levine]{fu2020d4rl}
Fu, J., Kumar, A., Nachum, O., Tucker, G., and Levine, S.
\newblock D4rl: Datasets for deep data-driven reinforcement learning.
\newblock \emph{arXiv preprint arXiv:2004.07219}, 2020.

\bibitem[Fujimoto \& Gu(2021)Fujimoto and Gu]{fujimoto2021minimalist}
Fujimoto, S. and Gu, S.~S.
\newblock A minimalist approach to offline reinforcement learning.
\newblock \emph{Advances in Neural Information Processing Systems}, 34, 2021.

\bibitem[Fujimoto et~al.(2019)Fujimoto, Meger, and Precup]{fujimoto2019off}
Fujimoto, S., Meger, D., and Precup, D.
\newblock Off-policy deep reinforcement learning without exploration.
\newblock In \emph{International Conference on Machine Learning}, pp.\
  2052--2062. PMLR, 2019.

\bibitem[Ghasemipour et~al.(2021)Ghasemipour, Schuurmans, and
  Gu]{ghasemipour2021emaq}
Ghasemipour, S. K.~S., Schuurmans, D., and Gu, S.~S.
\newblock Emaq: Expected-max q-learning operator for simple yet effective
  offline and online rl.
\newblock In \emph{International Conference on Machine Learning}, pp.\
  3682--3691. PMLR, 2021.

\bibitem[Ghosh et~al.(2019)Ghosh, Gupta, Reddy, Fu, Devin, Eysenbach, and
  Levine]{ghosh2019learning}
Ghosh, D., Gupta, A., Reddy, A., Fu, J., Devin, C., Eysenbach, B., and Levine,
  S.
\newblock Learning to reach goals via iterated supervised learning.
\newblock \emph{arXiv preprint arXiv:1912.06088}, 2019.

\bibitem[Guo et~al.(2021)Guo, Agrawal, Grover, Muthukumar, and
  Pananjady]{guo2021learning}
Guo, W., Agrawal, K.~K., Grover, A., Muthukumar, V., and Pananjady, A.
\newblock Learning from an exploring demonstrator: Optimal reward estimation
  for bandits.
\newblock \emph{arXiv preprint arXiv:2106.14866}, 2021.

\bibitem[Janner et~al.(2021)Janner, Li, and Levine]{janner2021offline}
Janner, M., Li, Q., and Levine, S.
\newblock Offline reinforcement learning as one big sequence modeling problem.
\newblock \emph{Advances in neural information processing systems}, 34, 2021.

\bibitem[Jaques et~al.(2019)Jaques, Ghandeharioun, Shen, Ferguson, Lapedriza,
  Jones, Gu, and Picard]{jaques2019way}
Jaques, N., Ghandeharioun, A., Shen, J.~H., Ferguson, C., Lapedriza, A., Jones,
  N., Gu, S., and Picard, R.
\newblock Way off-policy batch deep reinforcement learning of implicit human
  preferences in dialog.
\newblock \emph{arXiv preprint arXiv:1907.00456}, 2019.

\bibitem[Kidambi et~al.(2020)Kidambi, Rajeswaran, Netrapalli, and
  Joachims]{kidambi2020morel}
Kidambi, R., Rajeswaran, A., Netrapalli, P., and Joachims, T.
\newblock Morel: Model-based offline reinforcement learning.
\newblock \emph{Advances in neural information processing systems},
  33:\penalty0 21810--21823, 2020.

\bibitem[Kostrikov et~al.(2021{\natexlab{a}})Kostrikov, Fergus, Tompson, and
  Nachum]{kostrikov2021offline}
Kostrikov, I., Fergus, R., Tompson, J., and Nachum, O.
\newblock Offline reinforcement learning with fisher divergence critic
  regularization.
\newblock In \emph{International Conference on Machine Learning}, pp.\
  5774--5783. PMLR, 2021{\natexlab{a}}.

\bibitem[Kostrikov et~al.(2021{\natexlab{b}})Kostrikov, Nair, and Levine]{iql}
Kostrikov, I., Nair, A., and Levine, S.
\newblock Offline reinforcement learning with implicit q-learning.
\newblock \emph{arXiv preprint arXiv:2110.06169}, 2021{\natexlab{b}}.

\bibitem[Krishnamoorthy et~al.(2022)Krishnamoorthy, Mashkaria, and
  Grover]{krishnamoorthy2022generative}
Krishnamoorthy, S., Mashkaria, S.~M., and Grover, A.
\newblock Generative pretraining for black-box optimization.
\newblock \emph{arXiv preprint arXiv:2206.10786}, 2022.

\bibitem[Kumar \& Levine(2020)Kumar and Levine]{kumar2020model}
Kumar, A. and Levine, S.
\newblock Model inversion networks for model-based optimization.
\newblock \emph{Advances in Neural Information Processing Systems},
  33:\penalty0 5126--5137, 2020.

\bibitem[Kumar et~al.(2019)Kumar, Fu, Soh, Tucker, and
  Levine]{kumar2019stabilizing}
Kumar, A., Fu, J., Soh, M., Tucker, G., and Levine, S.
\newblock Stabilizing off-policy q-learning via bootstrapping error reduction.
\newblock \emph{Advances in Neural Information Processing Systems}, 32, 2019.

\bibitem[Kumar et~al.(2020)Kumar, Zhou, Tucker, and
  Levine]{kumar2020conservative}
Kumar, A., Zhou, A., Tucker, G., and Levine, S.
\newblock Conservative q-learning for offline reinforcement learning.
\newblock \emph{Advances in Neural Information Processing Systems},
  33:\penalty0 1179--1191, 2020.

\bibitem[Lambert et~al.(2022)Lambert, Wulfmeier, Whitney, Byravan, Bloesch,
  Dasagi, Hertweck, and Riedmiller]{lambert2022challenges}
Lambert, N., Wulfmeier, M., Whitney, W., Byravan, A., Bloesch, M., Dasagi, V.,
  Hertweck, T., and Riedmiller, M.
\newblock The challenges of exploration for offline reinforcement learning.
\newblock \emph{arXiv preprint arXiv:2201.11861}, 2022.

\bibitem[Lange et~al.(2012)Lange, Gabel, and Riedmiller]{lange2012batch}
Lange, S., Gabel, T., and Riedmiller, M.
\newblock Batch reinforcement learning.
\newblock In \emph{Reinforcement learning}, pp.\  45--73. Springer, 2012.

\bibitem[Laskin et~al.(2021)Laskin, Yarats, Liu, Lee, Zhan, Lu, Cang, Pinto,
  and Abbeel]{laskin2021urlb}
Laskin, M., Yarats, D., Liu, H., Lee, K., Zhan, A., Lu, K., Cang, C., Pinto,
  L., and Abbeel, P.
\newblock Urlb: Unsupervised reinforcement learning benchmark.
\newblock \emph{arXiv preprint arXiv:2110.15191}, 2021.

\bibitem[Levine et~al.(2020)Levine, Kumar, Tucker, and Fu]{levine2020offline}
Levine, S., Kumar, A., Tucker, G., and Fu, J.
\newblock Offline reinforcement learning: Tutorial, review, and perspectives on
  open problems.
\newblock \emph{arXiv preprint arXiv:2005.01643}, 2020.

\bibitem[Lu et~al.(2022)Lu, Grover, Abbeel, and Mordatch]{lu2021pretrained}
Lu, K., Grover, A., Abbeel, P., and Mordatch, I.
\newblock Pretrained transformers as universal computation engines.
\newblock In \emph{Proceedings of the AAAI Conference on Artificial
  Intelligence}, 2022.

\bibitem[Mnih et~al.(2015)Mnih, Kavukcuoglu, Silver, Rusu, Veness, Bellemare,
  Graves, Riedmiller, Fidjeland, Ostrovski, et~al.]{mnih2015human}
Mnih, V., Kavukcuoglu, K., Silver, D., Rusu, A.~A., Veness, J., Bellemare,
  M.~G., Graves, A., Riedmiller, M., Fidjeland, A.~K., Ostrovski, G., et~al.
\newblock Human-level control through deep reinforcement learning.
\newblock \emph{nature}, 518\penalty0 (7540):\penalty0 529--533, 2015.

\bibitem[Nachum et~al.(2019)Nachum, Dai, Kostrikov, Chow, Li, and
  Schuurmans]{nachum2019algaedice}
Nachum, O., Dai, B., Kostrikov, I., Chow, Y., Li, L., and Schuurmans, D.
\newblock Algaedice: Policy gradient from arbitrary experience.
\newblock \emph{arXiv preprint arXiv:1912.02074}, 2019.

\bibitem[Nguyen \& Grover(2022)Nguyen and Grover]{nguyen2022transformer}
Nguyen, T. and Grover, A.
\newblock Transformer neural processes: Uncertainty-aware meta learning via
  sequence modeling.
\newblock \emph{arXiv preprint arXiv:2207.04179}, 2022.

\bibitem[Radford et~al.(2018)Radford, Narasimhan, Salimans, and
  Sutskever]{radford2018improving}
Radford, A., Narasimhan, K., Salimans, T., and Sutskever, I.
\newblock Improving language understanding by generative pre-training.
\newblock 2018.

\bibitem[Reed et~al.(2022)Reed, Zolna, Parisotto, Colmenarejo, Novikov,
  Barth-Maron, Gimenez, Sulsky, Kay, Springenberg, et~al.]{reed2022generalist}
Reed, S., Zolna, K., Parisotto, E., Colmenarejo, S.~G., Novikov, A.,
  Barth-Maron, G., Gimenez, M., Sulsky, Y., Kay, J., Springenberg, J.~T.,
  et~al.
\newblock A generalist agent.
\newblock \emph{arXiv preprint arXiv:2205.06175}, 2022.

\bibitem[Schmidhuber(2019)]{schmidhuber2019reinforcement}
Schmidhuber, J.
\newblock Reinforcement learning upside down: Don't predict rewards--just map
  them to actions.
\newblock \emph{arXiv preprint arXiv:1912.02875}, 2019.

\bibitem[Srivastava et~al.(2019)Srivastava, Shyam, Mutz, Ja{\'s}kowski, and
  Schmidhuber]{srivastava2019training}
Srivastava, R.~K., Shyam, P., Mutz, F., Ja{\'s}kowski, W., and Schmidhuber, J.
\newblock Training agents using upside-down reinforcement learning.
\newblock \emph{arXiv preprint arXiv:1912.02877}, 2019.

\bibitem[Vaswani et~al.(2017)Vaswani, Shazeer, Parmar, Uszkoreit, Jones, Gomez,
  Kaiser, and Polosukhin]{vaswani2017attention}
Vaswani, A., Shazeer, N., Parmar, N., Uszkoreit, J., Jones, L., Gomez, A.~N.,
  Kaiser, {\L}., and Polosukhin, I.
\newblock Attention is all you need.
\newblock \emph{Advances in neural information processing systems}, 30, 2017.

\bibitem[Wang et~al.(2020)Wang, Foster, and Kakade]{wang2020statistical}
Wang, R., Foster, D.~P., and Kakade, S.~M.
\newblock What are the statistical limits of offline rl with linear function
  approximation?
\newblock \emph{arXiv preprint arXiv:2010.11895}, 2020.

\bibitem[Wu et~al.(2019)Wu, Tucker, and Nachum]{wu2019behavior}
Wu, Y., Tucker, G., and Nachum, O.
\newblock Behavior regularized offline reinforcement learning.
\newblock \emph{arXiv preprint arXiv:1911.11361}, 2019.

\bibitem[Yan et~al.(2021)Yan, Zhang, Abbeel, and Srinivas]{yan2021videogpt}
Yan, W., Zhang, Y., Abbeel, P., and Srinivas, A.
\newblock Videogpt: Video generation using vq-vae and transformers.
\newblock \emph{arXiv preprint arXiv:2104.10157}, 2021.

\bibitem[Yarats et~al.(2022)Yarats, Brandfonbrener, Liu, Laskin, Abbeel,
  Lazaric, and Pinto]{yarats2022don}
Yarats, D., Brandfonbrener, D., Liu, H., Laskin, M., Abbeel, P., Lazaric, A.,
  and Pinto, L.
\newblock Don't change the algorithm, change the data: Exploratory data for
  offline reinforcement learning.
\newblock \emph{arXiv preprint arXiv:2201.13425}, 2022.

\bibitem[Yu et~al.(2020)Yu, Thomas, Yu, Ermon, Zou, Levine, Finn, and
  Ma]{yu2020mopo}
Yu, T., Thomas, G., Yu, L., Ermon, S., Zou, J.~Y., Levine, S., Finn, C., and
  Ma, T.
\newblock Mopo: Model-based offline policy optimization.
\newblock \emph{Advances in Neural Information Processing Systems},
  33:\penalty0 14129--14142, 2020.

\bibitem[Zanette(2021)]{zanette2021exponential}
Zanette, A.
\newblock Exponential lower bounds for batch reinforcement learning: Batch rl
  can be exponentially harder than online rl.
\newblock In \emph{International Conference on Machine Learning}, pp.\
  12287--12297. PMLR, 2021.

\end{thebibliography}
\bibliographystyle{icml2023}

%%%%%%%%%%%%%%%%%%%%%%%%%%%%%%%%%%%%%%%%%%%%%%%%%%%%%%%%%%%%%%%%%%%%%%%%%%%%%%%
%%%%%%%%%%%%%%%%%%%%%%%%%%%%%%%%%%%%%%%%%%%%%%%%%%%%%%%%%%%%%%%%%%%%%%%%%%%%%%%
% APPENDIX
%%%%%%%%%%%%%%%%%%%%%%%%%%%%%%%%%%%%%%%%%%%%%%%%%%%%%%%%%%%%%%%%%%%%%%%%%%%%%%%
%%%%%%%%%%%%%%%%%%%%%%%%%%%%%%%%%%%%%%%%%%%%%%%%%%%%%%%%%%%%%%%%%%%%%%%%%%%%%%%
\newpage
\appendix
\onecolumn
\section{List of Symbols}
\vspace{-0.5cm}
\begin{table}[h]
    \small
    \centering
    \caption{Important symbols used in this paper.}
    \label{table:symbols}
    \begin{tabular}{l l l}
    \toprule
        Symbol &  Meaning & Definition \\
    \midrule
        $\S$ & state space \\
        $\A$ & action space \\
        $\tau$ & trajectory \\
        $|\tau|$ & trajectory length\\
        $\Tau$ & distribution of trajectories \\
        $\offlinedata$ & offline dataset  \\
        $\pi$ & policy \\
        $\theta$ & policy parameters \\
        $s_t$  & state at timestep $t$ \\
        $a_t$  & action at timestep $t$ \\
        $r_t$  & reward at timestep $t$ \\
        $r_\tau$ & trajectory return & $\sum_{t=1}^{|\tau|} r_t$\\
        $g_t$ & return-to-go at timestep $t$ & $\sum_{t'=t}^{|\tau|} r_t'$\\
        $H$ & maximum trajectory length \\
        $\omega_t$ & average return-to-go at timestep $t$ & $g_t / (H-t+1)$\\
        $\mathcal{L}_\text{RvS}$ & empirical risk of RvS & Equation~\eqref{eq:obj_rvs}\\
        $\mathcal{C}_\text{RvS}$ & conservative regularizer for RvS & Equation~\eqref{eq:conservatism_rvs}\\
        % $\alpha$ & coefficient to balance $\mathcal{L}_\text{RvS}$ and $\mathcal{C}_\text{RvS}$\\
        $f_{\Tau}(\tau)$ & probability density of trajectory $\tau \sim \Tau$ \\
        $p_{\TildeTau}(\tau)$ & probability density of trajectory $\tau \sim \TildeTau$ & Equation~\eqref{eq:transformed_traj_density}\\
        $b$ & index of a bin of trajectories in the offline dataset \\
        $|b|$ & size of bin $b$ \\
        $f_{\offlinedata}(b)$ & proportion of trajectories in bin $b$ & $|b| / |\offlinedata| $\\
        $\P_\text{bin}(b)$ & probability that bin $b$ is sampled & Equation~\eqref{eq:bin_sampling_probability}\\
        $\bar{r}^b_\tau$ & average return of trajectories in bin $b$ \\
        $\hat{r}^\star$ & highest return in the offline dataset \\
        $\hat{r}_q$ & $q$-th percentile of the returns in the offline dataset \\
    \bottomrule
    \end{tabular}
\end{table}

\vspace{-0.2cm}
\section{Implementation Details} \label{sec:appendix}
\subsection{Datasets and source code}

We train and evaluate our models on the D4RL~\cite{fu2020d4rl} and Atari~\citep{agarwal2020optimistic} benchmarks, which are available at \url{https://github.com/rail-berkeley/d4rl} and \url{https://research.google/tools/datasets/dqn-replay}, respectively.
% The two datasets are licensed under Creative Commons Attribution 4.0 License (CC BY), and the code is licensed under Apache 2.0 License.
Our codebase is largely based on the RvS~\cite{emmons2021rvs} official implementation at \url{https://github.com/scottemmons/rvs}, and DT~\cite{chen2021decision} official implementation at \url{https://github.com/kzl/decision-transformer}.
% Both codebases are licensed under MIT License.

% For all our experiments, each model was trained using 10 Intel(R) Xeon(R) CPU cores (E5-2698 v4 @ 2.20GHz) and one NVIDIA Tesla V100 SXM2 GPU. 

\newpage
\subsection{Default hyperparameters} \label{sec:hyper}

\begin{table}[th]
    \small
    \centering
    \caption{Hyperparameters used for locomotion experiments.}
    \label{table:hyper_gym}
    \begin{tabular}{@{}lll@{}}
\toprule
                                             & Hyperparameter               & Value                            \\ \midrule
\multirow{4}{*}{Model}                       & Context length $K$ (DT)           & $20$ \\
                                             & Number of attention heads (DT)   & $1$ \\
                                             & Hidden layers                & $2$ for RvS, $3$ for DT                             \\
                                             & Hidden dimension             & $1024$ for RvS, $128$ for DT                          \\
                                             & Activation function          & ReLU                             \\
                                             & Dropout                      & $0.0$ for RvS, $0.1$ for DT                           \\ 
                                                \midrule
\multirow{3}{*}{Conservative regularizer}    & Conservative percentile $q$  & $95$                             \\
                                             & Noise standard deviation $\sigma$           & $1000$               \\            
                                             & Regularization coefficient $\alpha$ & $1.0$                 \\ \midrule
\multirow{3}{*}{Trajectory weighting}        & \# bins $B$                  & $20$                             \\ 
                                             & Smoothing parameter $\lambda$                    & $0.01$                           \\
                                             & Smoothing parameter $\kappa$                     & $\hat{r}^{\star} - \hat{r}_{90}$ \\  \midrule
\multirow{4}{*}{Optimization}                & Batch size                   & $64$                             \\
                                             & Learning rate                & $1e-3$ for RvS, $1e-4$ for DT                         \\
                                             & Weight decay                 & $1e-4$
                                                            \\
                                             & Training iterations          & $100000$                         \\ 
                                              \midrule
Evaluation                                   & Target return                & $1 \times$ Expert return \\ \bottomrule
\end{tabular}
% }
\end{table}

\begin{table}[th]
    \small
    \centering
    \caption{Hyperparameters used for Atari experiments.}
    \label{table:rvs_hyper_atari}
    \begin{tabular}{@{}lll@{}}
\toprule
                                             & Hyperparameter               & Value                                                                                       \\ \midrule
\multirow{7}{*}{Model}                       & Encoder channels             & $32$, $32$, $64$                                                                            \\
                                             & Encoder filter sizes         & $8 \times 8$, $4 \times 4$, $3 \times 3$                                                    \\
                                             & Encoder strides              & $4$, $2$, $1$                                                                               \\
                                             & Hidden layers                & $4$                                                                                         \\
                                             & Hidden dimension             & $1024$                                                                                      \\
                                             & Activation function          & ReLU                                                                                        \\
                                             & Dropout                      & $0.1$                                                                                       \\ \midrule
\multirow{4}{*}{Conservative regularization} & Conservative percentile $q$  & $95$                                                                                        \\
                                             & Noise std $\sigma$           & \begin{tabular}[c]{@{}l@{}}$50$ for Breakout, Pong\\ $500$ for Qbert, Seaquest\end{tabular} \\
                                             & Conservative weight $\alpha$ & $0.1$                                                                                       \\ \midrule
\multirow{3}{*}{Trajectory weighting}        & \# bins $B$                  & $20$                                                                                        \\
                                             & $\lambda$                    & $0.1$                                                                                       \\
                                             & $\kappa$                     & $\hat{r}^{\star} - \hat{r}_{50}$                                                            \\ \midrule
\multirow{4}{*}{Optimization}                & Batch size                   & $128$                                                                                       \\
                                             & Learning rate                & $6e-4$                                                                                      \\
                                             & Weight decay                 & $1e-4$                                                                                      \\
                                             & Training iterations          & $25000$                                                                                     \\ \midrule 
Evaluation                 & Target return & 
                                            \begin{tabular}[c]{@{}l@{}}$90$ for Breakout ($1 \times$ max in dataset) \\
                                            $2500$ for Qbert ($5\times$ max in dataset) \\
                                            $20$ for Pong ($1\times$ max in dataset)                       \\
                                            $1450$ for Seaquest ($5\times$ max in dataset)                       \\
                                            \end{tabular} \\ \bottomrule
\end{tabular}
% }
\end{table}

\vspace{-0.2cm}
\newpage
\section{Ablation analysis}\label{app:ablation}
% For the experiments presented in Section~\ref{sec:expr}, we did not tune the hyperparameters of CWBC extensively, and the default values listed in Table~\ref{table:default_hyper} work well across different datasets. 
In this section, we investigate the impact of each of those hyperparameters on CWBC to give insights on what values work well in practice. We use the \walker environment and the three related datasets for illustration. In all the experiments, when we vary one hyperparameter, the other hyperparameters are kept as in Table~\ref{table:hyper_gym}.
% For easier inspection, we did not visualize the standard deviation, even though we aggregated the results over 10 seeds. For completeness, we plot figures with error bars in Section~\ref{sec:fig_with_error_bars}.

% \tn{for kappa and lambda, can show influence on the transformed distribution as well as performance}
% \tn{show results with different sets of hyperparameters for RvS, but say that we did not tune extensively and only used the default value in the main experiments}

\subsection{Trajectory Weighting: smoothing parameters $\lambda$ and $\kappa$} \label{sec:kappa_lambda}
Two hyperparameters $\kappa$ and $\lambda$ in Equation~\eqref{eq:bin_sampling_probability} affect the probability a bin index $b$ is sampled:
\[
    \P_{\text{bin}}(b) \ \propto \ \tfrac{f_{\offlinedata}(b)}{f_{\offlinedata}(b) + \lambda} \cdot \exp \big(- \tfrac{|\widebar{r}^b_{\tau}- \widehat{r}^{\star}|}{\kappa} \big).
\]
In practice, we have observed that the performance of CWBC is considerably robust to a wide range of values of $\kappa$ and $\lambda$.
% \qq{might need to be mellow here....even though i wrote this sentence, i'm not quite confident. It looks like there are some non-negligible differences.... i think we shouldn't fix the ylim for this paper. I used to suggest that because we were plotting different models per subplot....}
% It is slightly sensitive to the choice of $\kappa$ but stable across different choices of $\lambda$ we consider.

\begin{figure}[ht]
    \centering
    \includegraphics[width=0.9\textwidth]{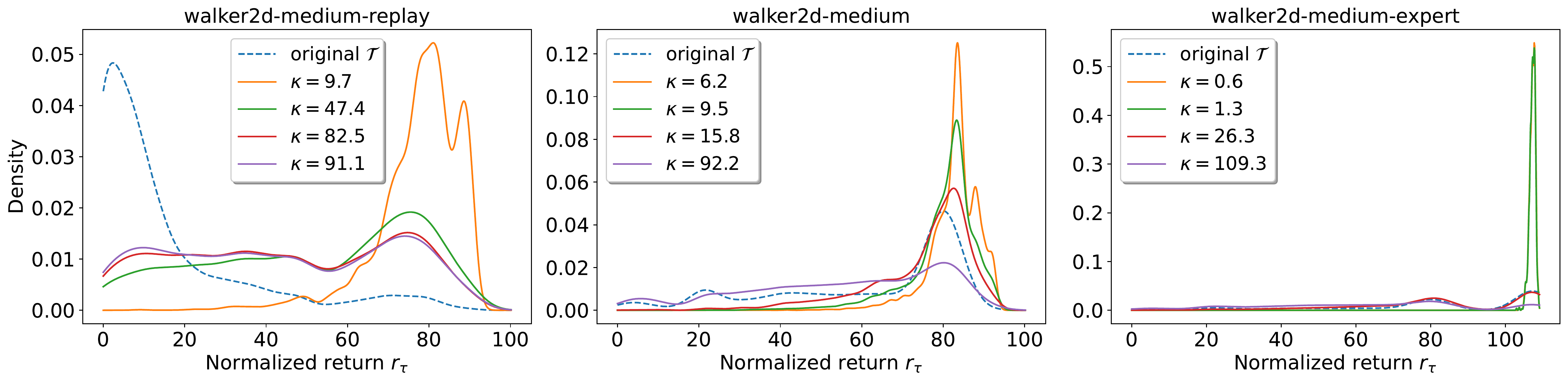}
    \includegraphics[width=0.9\textwidth]{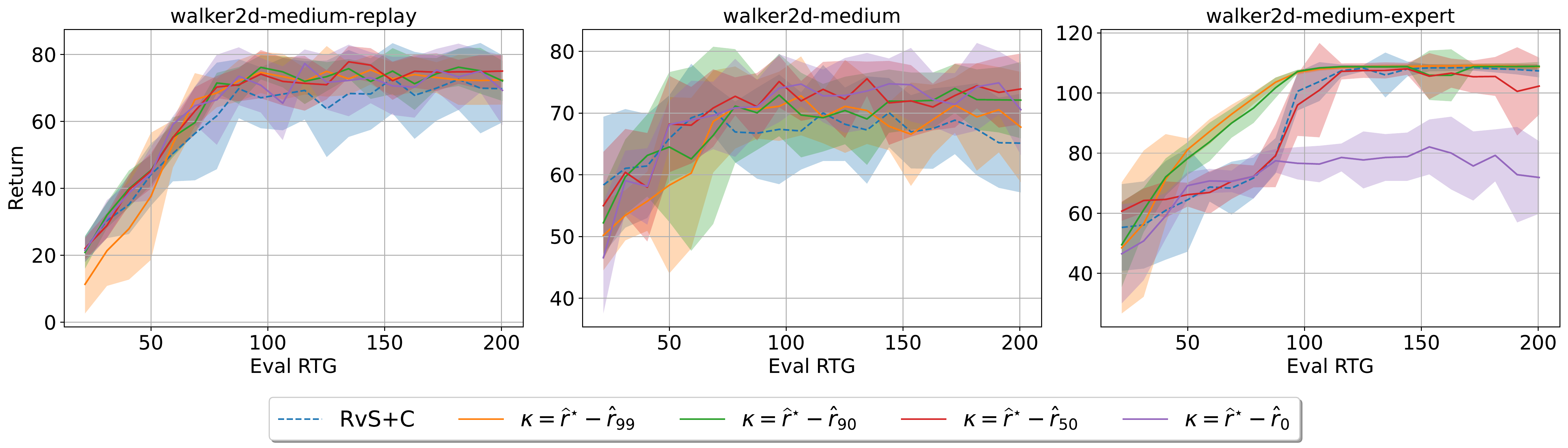}
    \caption{The influence of $\kappa$ on the transformed distribution (top) and on the performance of CWBC (bottom). The legend in each panel (top) shows the absolute values of $\kappa$ for easier comparison. In the bottom row, we also plot the results of RvS+C (no trajectory weighting) as a baseline.
    % Original means trajectory weighting is turned off.
    % \qq{I'd order kappas low to high, this is consistent with the text, also  commonly people will think as $\kappa$ increase what happens.... Also, original here is RvS + C, not the original RvS. Because we use "original RvS" in the main paper all the time, it's better to avoid reloading the term, I was confused why original performed so good here. Use a different legend for the bottom plot. Do not fix ylim to visualize the differences better. If it helps you can turn off variance. 
    %Maybe use a black dashed line for the original distribution and RvS+C. } \qq{I'll also order datasets \medreplay, \medium, \medexpert (low-to-high quality), this is for the discussion of $q$ below.}
    }
    \label{fig:tune_kappa}
\end{figure}

\paragraph{The impact of $\kappa$} The smoothing parameter $\kappa$ controls how we weight the trajectories based on their relative returns.
Intuitively, smaller $\kappa$ gives more weights to high-return bins (and thus their trajectories), and larger $\kappa$ makes the transformed distribution more uniform. We illustrate the effect of $\kappa$ on the transformed distribution and the performance of CWBC in Figure~\ref{fig:tune_kappa}. As in Section~\ref{sec:experiments}, we set $\kappa$ to be the difference between the empirical highest return $\hat{r}^\star$ and the $z$-th percentile return in the dataset: $\kappa = \hat{r}^\star - \hat{r}_z$,
and we vary the values of $z$. This allows the actual value of $\kappa$ to adapt to different datasets. 

Figure~\ref{fig:tune_kappa} shows the results. The top row plots the distributions of returns before and after trajectory weighting for varying values of $\kappa$. We tested four values $z \in \{99, 90, 50, 0\}$, which correspond to four increasing values of $\kappa$. We mark the actual values of $\kappa$ in each dataset in the top tow\footnote{$\hat{r}_0$ is defined to be the lowest return in the dataset: $\hat{r}_0 = \min_{\tau \in \offlinedata} r_\tau$.}. For each dataset, we can see the transformed distribution using small $\kappa$ (orange) highly concentrates on high returns; as $\kappa$ increases, the density for low returns increases and the distribution becomes more and more uniform. 
The bottom row plots the corresponding performance of CWBC with different choices of $\kappa$. We select RvS+C as our baseline model, which does not have trajectory weighting but has the conservative regularization enabled. We can see that relatively small values of $\kappa$ (based on $\hat{r}_{99}$ , $\hat{r}_{90}$ and $\hat{r}_{50}$) perform well on all the three datasets, whereas large values (based on $\hat{r}_{0}$) hurt the performance for the \medexpert dataset, and even underperform the baseline RvS+C.

% The performance of CWBC is stable for different values of $\kappa$?(\qq{might need to be mellow here....})

% \qq{do not mention trajectories for the plot; the plot only shows the returns' dentisty.}
% in the 
% Recall that in the main experiments, we set $\kappa$ to be the difference between the highest return and the $p=90$-th percentile return in the dataset, which allows for adaptive temperature for different datasets. Changing the value of $p$ will change the value of $\kappa$. 
% in which we vary the value of $p$.
% For easier comparison, we also compute the absolute values of $\kappa$, which are shown in the top row. 
% CWBC is not sensitive to the value of $\kappa$, only except for when $\kappa$ is very large ($\kappa = \hat{r}^{\star} - \hat{r}_{0}$). In \walker-\medexpert dataset, the transformed distribution corresponding to this particular value in (orange curve in the top row) has nearly zero density around the high-return trajectories, which explains the poor performance. In practice, we recommend visualizing the effect of $\kappa$ on the transformed distribution first, and pick the values which upweight the high-return trajectories.
% \begin{figure}[ht]
%     \centering
%     \includegraphics[width=0.95\textwidth]{figure/tune_kappa.pdf}
%     \caption{Performance of CWBC with different values of $\kappa$.}
%     \label{fig:tune_kappa}
% \end{figure}

\paragraph{The impact of $\lambda$} To better understand the role of $\lambda$, we can rewrite Equation~\eqref{eq:bin_sampling_probability} as
% the product of two terms T1 and T2 as follows:
\begin{equation*}
    \P_{\text{bin}}(b) \ \propto \ \overbrace{f_{\offlinedata}(b) \exp \big(- \tfrac{|\widebar{r}^b_{\tau}- \widehat{r}^{\star}|}{\kappa} \big) }^{\text{T1}} \cdot 
    \overbrace{ \set{ 1 / \big( f_{\offlinedata}(b) + \lambda \big)   } }^{\text{T2}}.
    \label{eq:bin_sampling_probability_form_2}
\end{equation*}
Clearly, only T2 depends on $\lambda$. 
When $\lambda = 0$, T2 is canceled out and the above equation reduces to
\[
   \P_{\text{bin}}(b) \ \propto \ \exp \big(- \tfrac{|\widebar{r}^b_{\tau}- \widehat{r}^{\star}|}{\kappa} \big),
\]
which purely depends on the relative return. As $\lambda$ increases, T2 is less sensitive to $f_{\offlinedata}(b)$,
and finally becomes the same for every $b \in [B]$
as $\lambda \rightarrow \infty$. In that scenario, $\P_{\text{bin}}(b)$ only depends on T1, which is the original frequency $f_{\offlinedata}(b)$ weighted by the relative return. 

\begin{figure}[ht]
    \centering
    \includegraphics[width=0.9\textwidth]{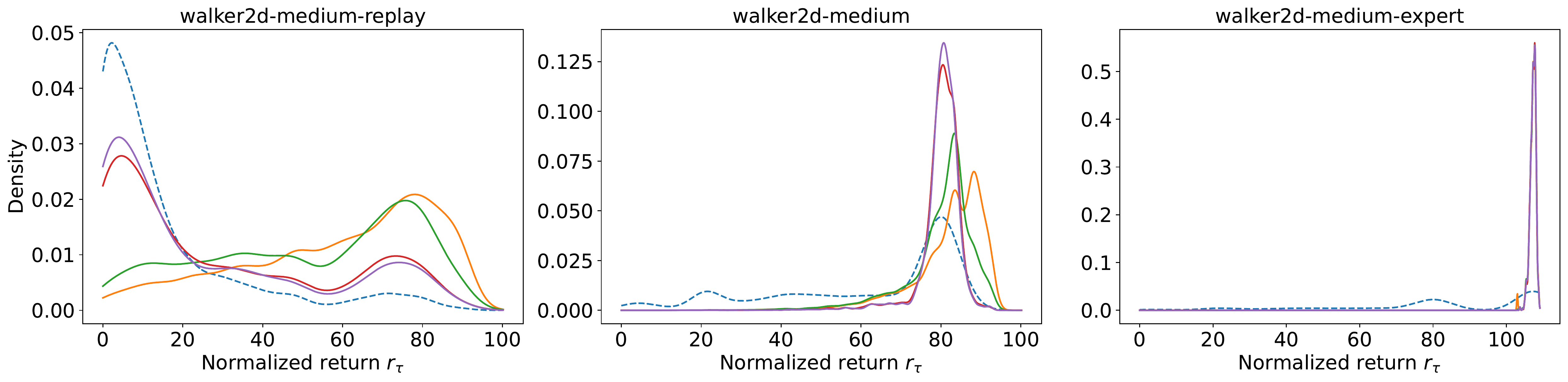}
    \includegraphics[width=0.9\textwidth]{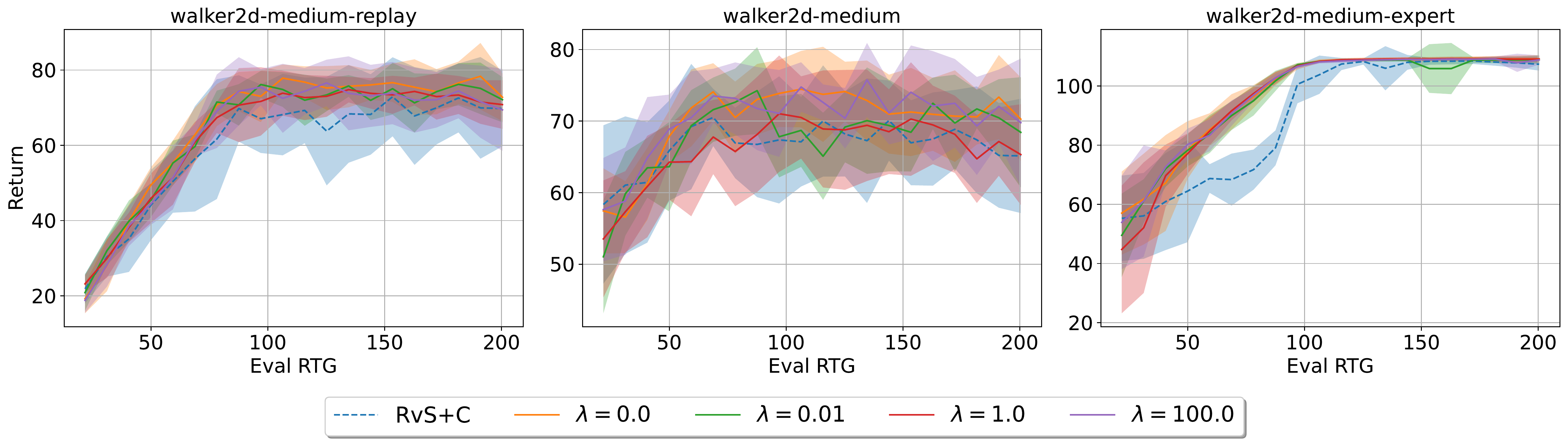}
    \caption{The influence of $\lambda$ on the transformed distribution (top) and on the performance of CWBC (bottom). We plot the result of RvS+C as the baseline. % For \walker\texttt{-}\medexpert, several curves overlap for returns greater than $100$.
    % \qq{Ditto for the original. include $\lambda=100$?}
    }
    \label{fig:tune_lambda}
\end{figure}

The top row of Figure~\ref{fig:tune_lambda} plots the distributions of returns before and after trajectory weighting with different values of $\lambda$. When $\lambda=0$, the distributions concentrate on high returns. As $\lambda$ increases, the distributions are more correlated with the original one, but still weights more on the high-return region compared to the original distribution due to the exponential term in T1. The bottom row of Figure~\ref{fig:tune_lambda} plots the actual performance of CWBC as $\lambda$ varies. All values of $\lambda$ produce similar results, which are consistently better than or comparable to training on the original datset (RvS+C).
% \qq{might need to change the ylim to visualize this better as well.}
% \qq{run for larger values of $\lambda$ }

% The hyperparameter $\lambda$ is used to weight the trajectories in Equation~\eqref{eq:bin_sampling_probability}. Intuitively, larger $\lambda$ means we give more weights to the bins whose original density is high, and smaller $\lambda$ means the original density has less impact on the transformed distribution.
% In other words, larger $\lambda$ encourages the transformed distribution $\TildeTau$ to stay close to the original data distribution $\Tau$.

% \qq{TODO: explain the transformation and exprs.}
% Figure~\ref{fig:tune_lambda} illustrates the influence of $\lambda$ on the shape of the transformed distribution as well as on the performance of CWBC. CWBC is not sensitive to the choice of $\lambda$, and shows consistently better results than training RvS on the original data distribution. The top row of Figure~\ref{fig:tune_lambda} shows that all values of $\lambda$ have the effect of upweighting high-return trajectories, leading to improved performance.
% vspace{-0.5cm}
% \begin{figure}[ht]
%     \centering
%     \includegraphics[width=0.95\textwidth]{figure/tune_lambda.pdf}
%     \caption{Performance of CWBC with different values of $\lambda$.}
%     \label{fig:tune_lambda}
% \end{figure}
% \subsection{Trajectory Weighting: temperature parameter $\kappa$}

\subsection{Conservative Regularization: percentile $q$}
\begin{figure}[H]
    \centering
    \includegraphics[width=0.9\textwidth]{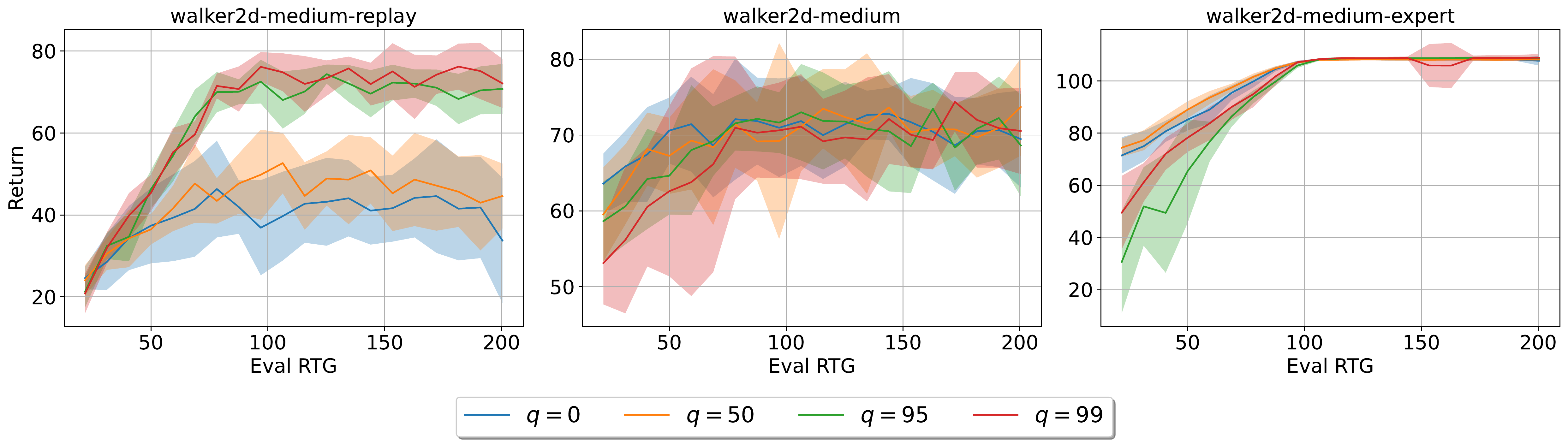}
    \caption{Performance of CWBC with different values of the conservative percentile $q$.
    % \qq{ylim as well}
    }
    \label{fig:tune_percentile}
\end{figure}

We only apply the conservative regularization to trajectories whose return is above the $q$-th percentile of the returns in the dataset. 
Intuitively, a larger $q$ applies the regularization to fewer trajectories. We test four values for $q$: $0, 50, 95$, and $99$. For $q=0$, our regularization applies to all the trajectories in the dataset.
Figure~\ref{fig:tune_percentile} demonstrates the impact of $q$ on the performance of CWBC.
$q=95$ and $q=99$ perform well on all the three datasets, while $q=50$ and $q=0$ lead to poor results for the \medreplay dataset.
This is because, when the regularization applies to trajectories of low returns,
% the values of noisy RTGs might be moderate,
% and
the regularizer will force the policy conditioned on out-of-distribution RTGs to stay close to the actions from low return trajectories. Since the \medreplay dataset contains many low return trajectories (see Figure~\ref{fig:tune_kappa}), such regularization results in poor performance. In contrast, \medium and \medexpert datasets contain a much larger portion of high return trajectories, and they are less 
sensitive to the choice of $q$.
% This is understandable, because when we condition on an out-of-distribution RTG, we want the policy to stay close to the chosen trajectories. If the low-return trajectories are chosen, then the policy is encouraged to stay close to these bad trajectories, resulting in poor performance. We note that this problem does not happen in \medium and \medexpert datasets, because even though the bad trajectories are chosen, the high-return trajectories are dominant in these datasets. Therefore, the policy will be biased towards the good trajectories, and will still perform well when $q$ is small. In practice, we recommend choosing a value of $q$ such that achieving $q$-th percentile return is sufficiently good for the problem of interest.

\subsection{Regularization Coefficient $\alpha$}
\begin{figure}[H]
    \centering
    \includegraphics[width=0.9\textwidth]{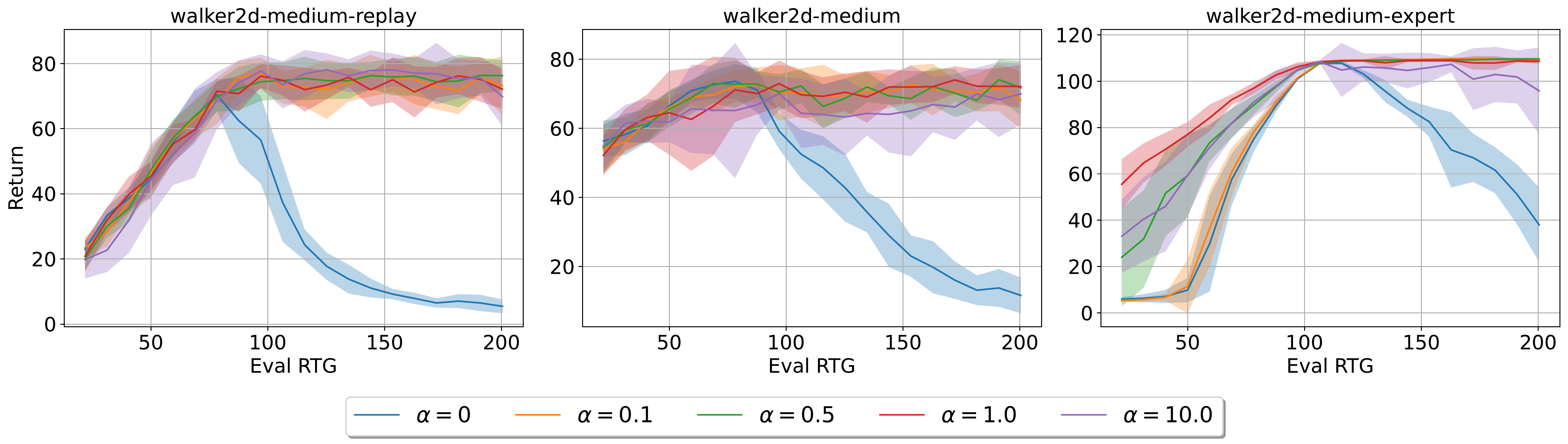}
    \caption{Performance of CWBC with different values of $\alpha$.}
    \label{fig:tune_alpha}
\end{figure}

The hyperparameter $\alpha$ controls the weight of the conservative regularization in the final objective function of CWBC $ \mathcal{L}_{\text{RvS}} + \alpha \cdot \mathcal{C}_{\text{RvS}}$. We show the performance of CWBC with different values of $\alpha$ in Figure~\ref{fig:tune_alpha}. Not using any regularization ($\alpha = 0$) suffers from the performance crash problem, while overly aggressive regularization ($\alpha = 10$) also hurts the performance. % , which we believe is due to the compromise of the prediction loss. 
CWBC is robust to the other non-extreme values of $\alpha$ .%, which all have good performance.

\section{Additional results on Atari games} \label{sec:atari_benchmark}

In addition to D4RL, we consider $4$ games from the Atari benchmark~\citep{bellemare2013arcade}: Breakout, Qbert, Pong, and Seaquest. Similar to~\citep{chen2021decision}, for each game, we train our method on $500000$ transitions sampled from the DQN-replay dataset, which consists of $50$ million transitions of an online DQN agent~\citep{mnih2015human}. Due to the varying performance of the DQN agent in different games, the quality of the datasets also varies. While Breakout and Pong datasets are high-quality with many expert transitions, Qbert and Seaquest datasets are highly suboptimal.
% This benchmark is challenging due to its high-dimensional visual inputs and delay between actions and rewards.

\textbf{Hyperparameters}
For trajectory weighting, we use $B = 20$ bins, $\lambda = 0.1$, and $\kappa = \wh{r}^{\star} - \wh{r}_{50}$. We apply conservative regularization with coefficient $\alpha = 0.1$ to trajectories whose returns are above $\wh{r}_{95}$. The standard deviation of the noise distribution varies across datasets, as each different games have very different return ranges. During evaluation, we set the target return to $5 \times \wh{r}^{\star}$ for Qbert and Seaquest, and to $1 \times \wh{r}^{\star}$ for Breakout and Pong.

\begin{table*}[ht]
    \centering
    \caption{Comparison of the normalized return on Atari games. The results are averaged over $3$ seeds. We include the results of DT, CQL, and BC from~\citep{chen2021decision} for reference.}
    \label{table:atari_results}
    \resizebox{1.0\columnwidth}{!}{
    \begin{tabular}{l || l l l l l l l}
    \toprule
    & RvS & RvS+W & RvS+C & RvS+W+C & DT & CQL & BC \\ \midrule
    Breakout & $126.9 \pm 38.0$ & $120.1 \pm 28.43$ & \by{$163.0 \pm 50.4$} & \by{$237.3 \pm 82.1$} & $267.5 \pm 97.5$ & $211.1$ & $138.9 \pm 61.7$ \\
    Qbert & $-0.4 \pm 0.2$ & $0.0 \pm 0.4$ & \by{$12.4 \pm 8.6$} & \by{$19.1 \pm 2.7$} & $15.4 \pm 11.4$ & $104.2$ & $17.3 \pm 14.7$ \\
    Pong & $75.7 \pm 8.6$ & \by{$90.7 \pm 6.4$} & \by{$84.1 \pm 9.6$} & \by{$90.4 \pm 1.9$} & $106.1 \pm 8.1$ & $111.9$ & $85.2 \pm 20.0$ \\
    Seaquest & $0.2 \pm 0.2$ & $-0.1 \pm 0.1$ & \by{$1.6 \pm 0.2$} & \by{$1.4 \pm 0.3$} & $2.5 \pm 0.4$ & $1.7$ & $2.1 \pm 0.3$ \\
    \midrule
    \# wins & / & $1$ & $4$ & $4$ \\
    average & $50.6$ & $52.7$ & $62.3$ & $87.1$ & $97.9$ & $107.2$ & $60.9$ \\
    \bottomrule
    \end{tabular}
    }
\end{table*}

\begin{figure}[ht]
    \centering
    \includegraphics[width=0.99\textwidth]{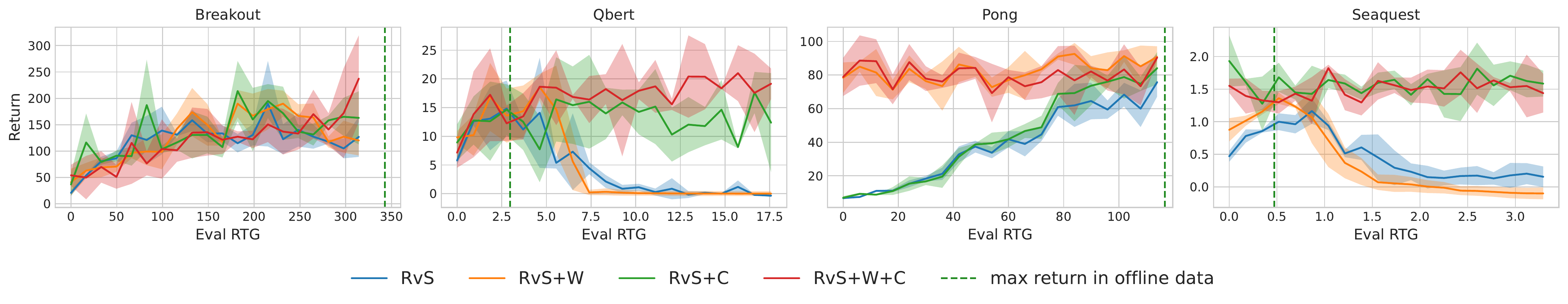}
    \caption{Performance of RvS and its variants on Atari games when conditioning on different evaluation RTGs.}
    \label{fig:eval_rvs_atari}
\end{figure}

\textbf{Results}
Table~\ref{table:atari_results} summarizes the performance of RvS and its variants. CWBC (RvS+W+C) is the best method, outperforming the original RvS by $72\%$ on average. Figure~\ref{fig:eval_rvs_atari} clearly shows the effectiveness of the conservative regularization (+C). In two low-quality datasets Qbert and Seaquest, the performance of RvS degrades quickly when conditioning on out-of-distribution RTGs. By regularizing the policy to stay close to the data distribution, we achieve a much more stable performance. The trajectory weighting component (+W) alone has varying effects on performance because of the performance crash problem, but achieves state-of-the-art when used in conjunction with conservative regularization.

It is also worth noting that in both Qbert and Seaquest, CWBC achieves returns that are much higher than the best return in the offline dataset. This shows that while conservatism encourages the policy to stay close to the data distribution, it does not prohibit extrapolation. There is always a trade-off between optimizing the original supervised objective (which presumably allows extrapolation) and the conservative objective. This is very similar to other conservative regularizations used in value-based such as CQL or TD3+BC, where there is a trade-off between learning the value function and staying close to the data distribution.

\section{Additional results on D4RL Antmaze} \label{sec:antmaze_benchmark}
Our proposed conservative regularization is especially important in dense reward environments such as gym locomotion tasks or Atari games, where choosing the target return during evaluation is a difficult problem. On the other hand, trajectory weighting is generally useful whenever the offline dataset contains both low-return and high-return trajectories. In this section, we consider Antmaze~\citep{fu2020d4rl}, a sparse reward environment in the D4RL benchmark to evaluate the generality of CWBC. Antmaze is a navigation domain in which the task is to control a complex 8-DoF "Ant" quadruped robot to reach a goal location. We consider $3$ maze layouts: \umaze, \medium, and \mazelarge, and $3$  dataset flavors: \vzero, \diverse, and \play. We use the same set of hyperparameters as mentioned in~\ref{sec:hyper}.
% 1) \vzero with a fixed goal and a fixed start point, 2) \diverse with randomized goals and randomized start locations, and 3) \play with a set of hand-picked start locations and a different set of hand-picked locations, which can be different from the goal at evaluation. 

\begin{table*}[h]
    \centering
    \caption{Comparison of the success rate on the Antmaze environment. The results are averaged over $3$ seeds. We include the results of DT, CQL, and BC from~\citep{emmons2021rvs} for reference.}
    \label{table:antmaze}
    \resizebox{1.0\columnwidth}{!}{
    % \begin{tabular}{l || l  l  l | l  l  l  | l  l  l | l | l}
    \begin{tabular}{l || l l l l l l l}
    \toprule
    & RvS & RvS+W & RvS+C & RvS+W+C & DT & CQL & BC \\ \midrule
    \umaze-\vzero & $54.0 \pm 13.56$ & \by{$65.0 \pm 18.03$} & \by{$58.0 \pm 8.72$} & \by{$65.0 \pm 12.85$} & $65.6$ & $44.8$ & $54.6$ \\
    \umaze-\diverse & $55.0 \pm 15.65$ & $46.0 \pm 16.85$ & $50.0 \pm 10.95$ & $42.0 \pm 7.48$ & $51.2$ & $23.4$ & $45.6$  \\ \midrule
    \mazemedium-\play & $0.0 \pm 0.0$ & \by{$26.0 \pm 12.0$} & $0.0 \pm 0.0$ & \by{$25.0 \pm 13.6$} & $1.0$ & $0.0$ & $0.0$ \\
    \mazemedium-\diverse & $1.0 \pm 3.0$ & \by{$24.0 \pm 15.62$} & $1.0 \pm 3.0$ & \by{$23.0 \pm 11.0$} & $0.6$ & $0.0$ & $0.0$ \\ \midrule
    \mazelarge-\play & $0.0 \pm 0.0$ & \by{$4.0 \pm 4.9$} & $0.0 \pm 0.0$ & \by{$5.0 \pm 6.71$} & $0.0$ & $0.0$ & $0.0$ \\
    \mazelarge-\diverse & $0.0 \pm 0.0$ & \by{$10.0 \pm 10.0$} & $0.0 \pm 0.0$ & \by{$17.0 \pm 11.87$} & $0.2$ & $0.0$ & $0.0$ \\ \midrule
    \# wins & / & $4$ & $1$ & $4$ \\
    average & $18.3$ & $29.2$ & $18.2$ & $29.5$ & $19.8$ & $11.4$ & $16.7$ \\
    \bottomrule
    \end{tabular}
    }
\end{table*}

\textbf{Results}
Table~\ref{table:antmaze} summarizes the results. As expected, the conservative regularization is not important in these tasks, as the target return is either 0 (fail) or 1 (success). However, the trajectory weighting significantly boosts performance, resulting in an average of $60\%$ improvement over the original RvS.

\section{Trajectory weighting versus Hard filtering}
An alternative to trajectory weighting is hard filtering (+F), where we train the model on only top $10\%$ trajectories with the highest returns. Filtering can be considered a hard weighting mechanism, wherein the transformed distribution only has support over trajectories with returns above a certain threshold.

\subsection{Hard filtering for RvS}
\begin{figure}[h]
    \centering
    \includegraphics[width=0.9\textwidth]{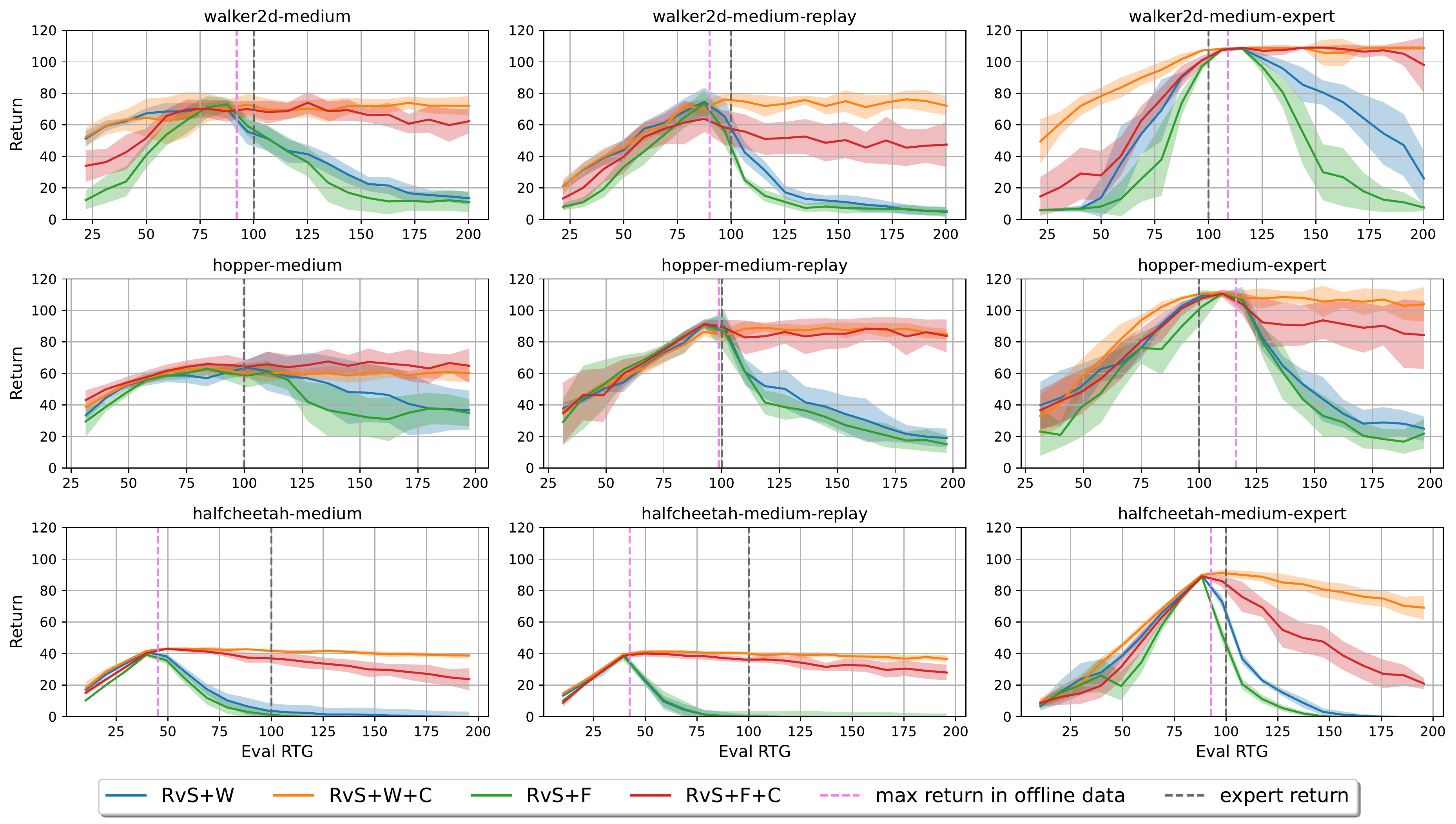}
    \caption{Comparison of trajectory weighting and hard filtering.}
    \label{fig:percent_rvs}
\end{figure}

When using hard filtering for RvS, we also consider combining it with the conservative regularization. Table~\ref{table:hard_filtering} and Figure~\ref{fig:percent_rvs} compare the performance of trajectory weighting and hard filtering when applied to RvS. While RvS+F+C also gains notable improvements , it lags behind RvS+W+C and seems to erode the benefits of conservatism alone in RvS+C. This agrees with our analysis in Section~\ref{subsec:algo_reweight}. While hard filtering achieves the same effect of reducing bias, it completely removes the low-return trajectories, resulting in highly increased variance. Our trajectory weighting upweights the good trajectories but aims to stay close to the original data distribution, balancing this bias-variance tradeoff. This is clearly shown in Figure~\ref{fig:percent_rvs}, where RvS+W+C has much smaller variance when conditioning on large RTGs.

\begin{table*}[h]
    \centering
    \caption{Comparison of trajectory weighting (+W) and hard filtering (+F) on D4RL locomotion benchmarks. The results are averaged over 10 seeds.
    }
    \label{table:hard_filtering}
    \resizebox{\columnwidth}{!}{
    \begin{tabular}{l || l l l l l l}
    \toprule
    & RvS & RvS+W & RvS+C & RvS+W+C & RvS+F & RvS+F+C  \\ \midrule
    \walker-\medium & $73.3 \pm 5.7$ & $54.5 \pm 7.7$ & $71.3 \pm 4.9$ & \by{$73.6 \pm 5.4$} & $60.9 \pm 4.9$ & $68.2 \pm 7.1$ \\
    \walker-\medreplay & $54.0 \pm 12.1$ & \by{$61.2 \pm 14.7$} & \by{$62.0 \pm 13.5$} & \by{$72.8 \pm 7.5$} & $47.1 \pm 7.7$ & $53.9 \pm 11.0$ \\
    \walker-\medexpert & $102.2 \pm 2.3$ & \by{$104.1 \pm 0.5$} & $102.1 \pm 10.2$ & \by{$107.6 \pm 0.5$} & $101.7 \pm 3.3$ & \by{$105.4 \pm 0.6$}  \\ \midrule
    \hopper-\medium & $56.6 \pm 5.5$ & \by{$62.5 \pm 7.1$} & \by{$61.0 \pm 5.3$} & \by{$62.9 \pm 3.6$} & \by{$62.4 \pm 5.0$} & \by{$65.7 \pm 6.4$}  \\
    \hopper-\medreplay & $87.7 \pm 9.7$ & \by{$92.4 \pm 6.1$} & \by{$91.5 \pm 3.5$} & \by{$87.7 \pm 4.2$} & \by{$91.2 \pm 5.3$} & \by{$92.1 \pm 2.9$} \\
    \hopper-\medexpert & $108.8 \pm 0.9$ & $108.4 \pm 1.8$ & $101.0 \pm 13.4$ & \by{$110.0 \pm 2.8$} & $97.5 \pm 15.0$ & $105.8 \pm 3.5$ \\ \midrule
    \cheetah-\medium & $16.2 \pm 4.5$ & $4.0 \pm 5.4$ & \by{$40.7 \pm 1.0$} & \by{$42.2 \pm 0.7$} & $1.4 \pm 3.3$ & \by{$36.2 \pm 2.5$}  \\
    \cheetah-\medreplay & $-0.4 \pm 2.7$ & $-0.8 \pm 2.2$ & \by{$36.8 \pm 1.5$} & \by{$40.4 \pm 0.8$} & \by{$-0.1 \pm 3.5$} & \by{$35.7 \pm 2.8$} \\
    \cheetah-\medexpert & $83.4 \pm 2.1$ & $69.1 \pm 3.7$ & \by{$91.2 \pm 1.0$} & \by{$91.1 \pm 2.0$} & $46.0 \pm 1.5$ & $83.2 \pm 5.0$ \\
    \midrule
    \# wins & / & $4$ & $6$ & \colorbox{Orange1}{$9$} & $3$ & $5$ \\
    average & $64.6$ & $61.7$ & $73.1$ & \colorbox{Orange1}{$76.5$} & $56.5$ & $71.8$  \\
    \bottomrule
    \end{tabular}
}
\end{table*}

\subsection{Hard filtering for unconditional BC}
Hard filtering can also be applied to ordinary BC. This is equivalent to Filtered BC in~\citep{emmons2021rvs}. Table~\ref{table:filtered_bc} compares Filtered BC and CWBC. CWBC performs comparably well in \medium and \medexpert datasets, and outperforms Filtered BC significantly with an average improvement of $12\%$ in \medreplay datasets. We believe that in low-quality datasets, even when we filter out $90\%$ percent of the data, the quality of the remaining trajectories is still very diverse that simple imitation learning is not good enough. CWBC is able to learn from such diverse data, and by conditioning on expert return at test time, we can recover an efficient policy.

\begin{table*}[h!]
    \centering
    \caption{The normalized return on D4RL for Filtered BC, RvS, and CWBC. For Filtered BC, we get the numbers from~\cite{emmons2021rvs}.}
    \label{table:filtered_bc}
    %\resizebox{0.5\columnwidth}{!}{
    \begin{tabular}{l || l l l}
    \toprule
    & Filtered BC & RvS & RvS+W+C \\ \midrule
    \walker-\medium & $75.0$ & $73.3 \pm 5.7$ & $73.6 \pm 5.4$ \\
    \hopper-\medium & $56.9$ & $56.6 \pm 5.5$ & \by{$62.9 \pm 3.6$} \\
    \cheetah-\medium & $42.5$ & $16.2 \pm 4.5$ & $42.2 \pm 0.7$ \\ \midrule
    \medium average & $58.1$ & $48.7$ & \by{$59.6$} \\ \midrule
    \walker-\medreplay & $62.5$ & $54.0 \pm 12.1$ & \by{$72.8 \pm 7.5$}  \\
    \hopper-\medreplay & $75.9$ & $87.7 \pm 9.7$ & \by{$87.7 \pm 5.2$} \\
    \cheetah-\medreplay & $40.6$ & $-0.4 \pm 2.7$ & $40.4 \pm 0.8$ \\ \midrule
    \medreplay average & $59.7$ & $47.1$ & \by{$67.0$} \\ \midrule
    \walker-\medexpert & $109.0$ & $102.2 \pm 2.3$ & $107.6 \pm 0.5$  \\
    \hopper-\medexpert & $110.9$ & $108.8 \pm 0.9$ & $110.0 \pm 2.8$ \\
    \cheetah-\medexpert & $92.9 \pm$ & $83.4 \pm 2.1$ & $91.1 \pm 2.0$ \\ \midrule
    \medexpert average & $104.3$ & $98.1$ & $102.9$ \\ \midrule
    average & $74.0$ & $64.6$ & \by{$76.5$} \\
    \bottomrule
    \end{tabular}
    %}
\end{table*}

\newpage
\section{Bias-variance tradeoff analysis} \label{sec:bias_variance}
We formalize our discussion on the bias-variance tradeoff when learning from a suboptimal distribution mentioned in Section~\ref{subsec:algo_reweight}. The objective functions for training DT (\ref{eq:dt_loss}) and RvS (\ref{eq:obj_rvs}) can be rewritten as:
\begin{align}
    \min_\theta \mathcal{L}_{p_\mathcal{D}}(\theta) & = \mathbb{E}_{\tau \sim \Tau} \left[D(\tau, \pi_\theta) \right] \\
    & = \mathbb{E}_{r \sim p_{\mathcal{D}}(r), \tau \sim \Tau_r} \left[D(\tau, \pi_\theta) \right]. \label{eq:rewritten_loss}
\end{align}
In which, $p_{\mathcal{D}}(r)$ is the data distribution over trajectory returns, $\Tau_r$ is a uniform distribution over the set of trajectories whose return is $r$, and $D(\tau, \pi_\theta)$ is the supervised loss function with respect to the sampled trajectory $\tau$. For DT, $D(\tau, \pi_\theta) = \textstyle \tfrac{1}{|\tau|}\sum_{t=1}^{|\tau|} \displaystyle \big(a_t - \pi_\theta(g_{t-K:t}, s_{t-K:t}, a_{t-K:t-1}) \big)^2$, and for RvS, $D(\tau, \pi_\theta) = \textstyle \tfrac{1}{|\tau|}\sum_{t=1}^{|\tau|} \displaystyle \big(a_t - \pi_\theta(s_t, \omega_t) \big)^2$. Equation (\ref{eq:rewritten_loss}) is equivalent to first sampling a return $r$, then sampling a trajectory $\tau$ whose return is $r$, and calculating the loss on $\tau$. Ideally, we want to train the model from an optimal return distribution $p^{\star}(r)$, which is centered around the expert return $\rstar$:
\begin{equation}
    \min_\theta \mathcal{L}_{p^\star}(\theta) = \mathbb{E}_{r \sim p^\star(r), \tau \sim \Tau_r}\left[D(\tau, \pi_\theta) \right].
\end{equation}
In practice, we only have access to the suboptimal return distribution $p_\mathcal{D}(r)$, which leads to a biased training objective with respect to $p^\star(r)$. While the dataset is fixed, we can transform the data distribution $p_\mathcal{D}(r)$ to $q(r)$ that better estimates the ideal distribution $p^\star(r)$. The objective function with respect to $q$ is:
\begin{align}
    \min_\theta \mathcal{L}_{q}(\theta) & = \mathbb{E}_{r \sim q(r), \tau \sim \Tau_r}\left[D(\tau, \pi_\theta) \right] \\
    & = \mathbb{E}_{r \sim p_\mathcal{D}(r), \tau \sim \Tau_r}\left[ \frac{q(r)}{p_\mathcal{D}(r)} \cdot D(\tau, \pi_\theta) \right]
\end{align}
In the extreme case, $q(r) = \mathbbm{1}[r = \rstar]$, which means we only train the policy on trajectories whose return matches the expert return $\rstar$. However, since offline datasets often contain very few expert trajectories, this $q$ leads to a very high-variance training objective. An optimal distribution $q$ should lead to a training objective that balances the bias-variance tradeoff. We quantify this by measuring the $\ell_2$ of the difference between the gradient of $\mathcal{L}_q(\theta)$ and the gradient of the optimal objective function $\mathcal{L}_{p^{\star}}(\theta)$. Analogous to~\citet{kumar2020model}, we can prove that for some constants $C_1, C_2, C_3$, with high confidence:
\begin{equation}
    \mathbb{E}\left[|| \nabla_\theta \mathcal{L}_q(\theta) - \nabla_\theta \mathcal{L}_{p^\star}(\theta) ||_2^2 \right] \leq C_1 \cdot \mathbb{E}_{r \sim q(r)} \left[ \frac{1}{N_r} \right] + C_2 \cdot \frac{d_2(q || p_\mathcal{D})}{|\mathcal{D}|} + C_3 \cdot D_{\text{TV}}(p^\star, q)^2. \label{eq:bias_variance}
\end{equation}
In which, $N_r$ is the number of trajectories in dataset $\mathcal{D}$ whose return is $r$, $d_2$ is the exponentiated Renyi divergence, and $D_{\text{TV}}$ is the total variation divergence. The right hand side of inequality (\ref{eq:bias_variance}) shows that an optimal distribution $q$ should be close to the data distribution $p_\mathcal{D}$ to reduce variance, while approximating well $p^\star$ to reduce bias. As shown in~\citet{kumar2020model}, $q(r) \propto \frac{N_r}{N_r + K} \cdot \exp(-\frac{|r - \rstar|}{\kappa})$ minimizes this bound, which inspires our trajectory weighting.

\end{document}